\documentclass[11pt]{article}

\usepackage[preprint]{acl}

\usepackage{times}
\usepackage{latexsym}
\usepackage{cuted}
\usepackage[T1]{fontenc}
\usepackage[utf8]{inputenc}

\usepackage{microtype}

\usepackage{inconsolata}

\usepackage{graphicx}

\usepackage{arydshln}
\usepackage{lipsum} 
\usepackage{tabularray}
\usepackage{multirow}
\usepackage{booktabs, caption, array}
\usepackage{algorithm}
\usepackage{algorithmic}
\usepackage{comment}
\usepackage{xcolor}
\usepackage{subcaption}
\usepackage{graphicx}
\usepackage{subcaption}
\usepackage{enumitem}
\usepackage{framed}
\usepackage{hyperref} 
\usepackage{amsmath}
\definecolor{shadecolor}{gray}{0.9}
\usepackage{comment}
\usepackage{xspace}
\usepackage{colortbl}
\usepackage{subcaption}
\usepackage{xcolor}
\usepackage{amssymb}   % 或 amsfonts
\usepackage{booktabs,multirow,tabularx,colortbl,xcolor}
\usepackage{tcolorbox}
\tcbuselibrary{breakable}
\newtcolorbox{promptbox}{
    colback=gray!15,
    colframe=gray!15,
    boxrule=0pt,
    arc=0pt,
    left=6pt, right=6pt, top=6pt, bottom=6pt,
    boxsep=0pt,
    fontupper=\raggedright\setlength{\parskip}{1em}\setlength{\parindent}{0pt}, 
    before skip=8pt,
    after skip=8pt,
    breakable
}
\definecolor{fppsh}{HTML}{E8F3FF}  % FPPS-H (light blue)
\definecolor{fppss}{HTML}{EAF7EE}  % FPPS-S (light green)
\definecolor{fppsm}{HTML}{FFF4E6}  % FPPS-M (light orange)
\definecolor{baseline}{HTML}{F7F7F7}
\title{When Personalization Misleads: Understanding and Mitigating Hallucinations in Personalized LLMs}

\author{
    Zhongxiang Sun\textsuperscript{1},
    Yi Zhan\textsuperscript{1},
    Chenglei Shen\textsuperscript{1},
    Weijie Yu\textsuperscript{3},
    Xiao Zhang\textsuperscript{1},
    Ming He\textsuperscript{2},
    Jun Xu\textsuperscript{1}\\
    \textsuperscript{1}Gaoling School of Artificial Intelligence, Renmin University of China; 
    \textsuperscript{2}AI Lab at Lenovo Research; \\
    \textsuperscript{3}School of Artificial Intelligence and Data Science, University of International Business and Economics.\\
    \texttt{\{sunzhongxiang\}@ruc.edu.cn}
}

\begin{document}
\maketitle
\begin{abstract}
Personalized large language models (LLMs) adapt model behavior to individual users to enhance user satisfaction, yet personalization can inadvertently distort factual reasoning. We show that when personalized LLMs face factual queries, there exists a phenomenon where the model generates answers aligned with a user's prior history rather than the objective truth, resulting in \textbf{personalization-induced hallucinations} that degrade factual reliability and may propagate incorrect beliefs, due to representational entanglement between personalization and factual representations.
To address this issue, we propose \textbf{Factuality-Preserving Personalized Steering (FPPS)}, a lightweight inference-time approach that mitigates personalization-induced factual distortions while preserving personalized behavior. We further introduce \textbf{PFQABench}, the first benchmark designed to jointly evaluate factual and personalized question answering under personalization. Experiments across multiple LLM backbones and personalization methods show that FPPS substantially improves factual accuracy while maintaining personalized performance.
\end{abstract}

 % To address this issue, we propose \textbf{Factuality-Preserving Personalized Steering (FPPS)}, a lightweight inference-time method that mitigates personalization-induced distortions using a \textbf{Representation Shift Locator} that identifies personalization-sensitive regions in the model, a \textbf{Factuality Entanglement Prober} that measures factual distortion, and an \textbf{Adaptive Knowledge Steering Module} that restores factual reasoning without compromising personalization. We further introduce \textbf{PFQABench}, the first benchmark for evaluating factual hallucination under personalization. Experiments across multiple LLM backbones and personalization methods show that FPPS substantially improves factual accuracy while preserving personalized performance.

\section{Introduction}
Personalized Large Language Models (LLMs) are increasingly deployed in real-world applications to adapt model behavior to individual users through mechanisms such as long-term memory, preference profiles, and historical interaction modeling. It has already become a \textbf{core product paradigm} in leading LLM systems~\citep{zhang2024personalized,google_gemini_personalization_2024,openai_chatgpt_memory_2024,claude_memory_2025}. In practice, such personalization features are often \textbf{enabled by default} and tightly integrated into the user experience, as they are widely regarded as a key driver of user engagement and retention in commercial LLM deployments~\cite{king2025user,geminis_new_personal_context_is_not_just_better}.

\begin{figure}[!t]
    \centering
    \includegraphics[width=\columnwidth]{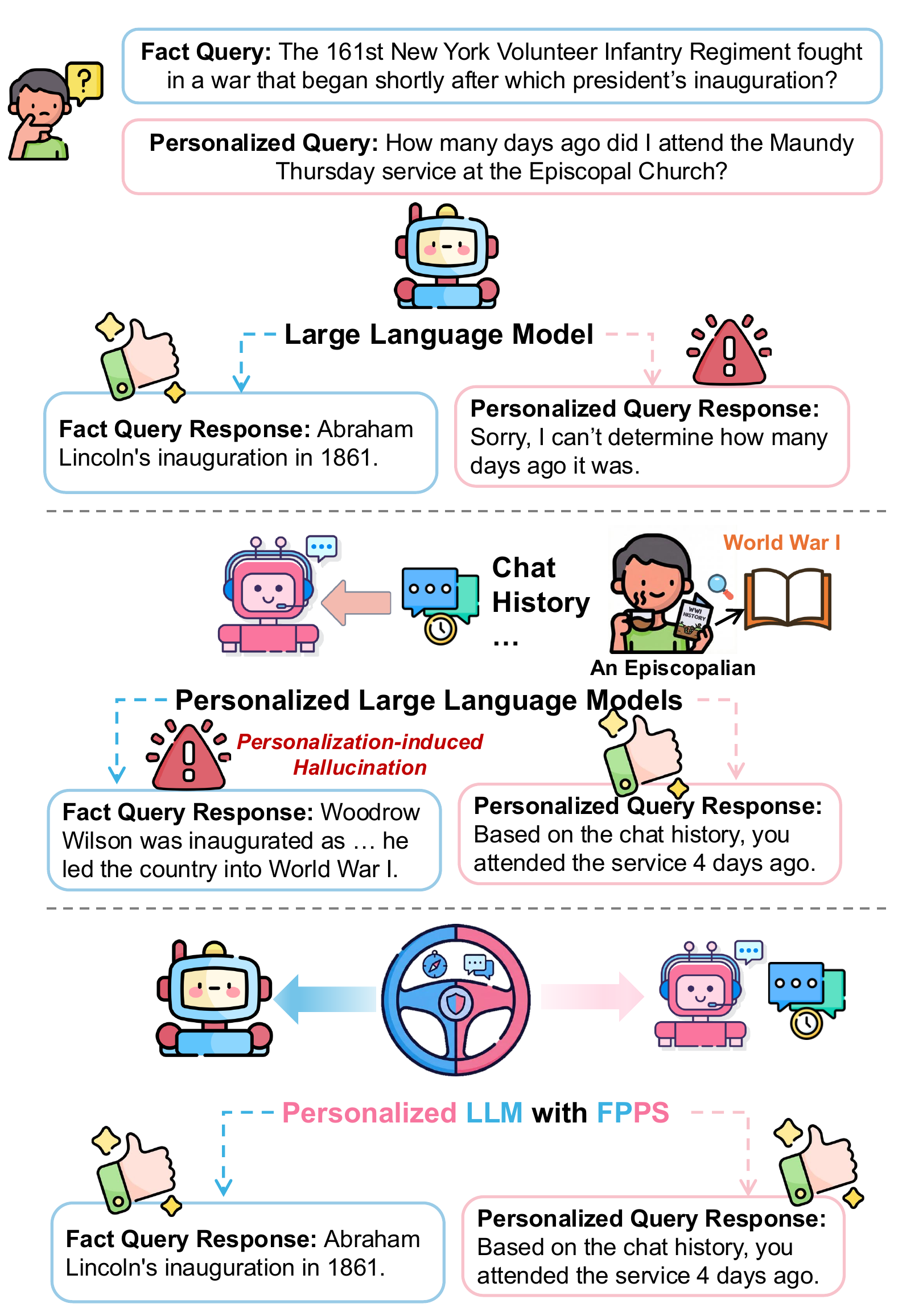}
    \caption{\textbf{Illustration of personalization-induced hallucinations and the effect of FPPS, using real examples obtained from our PFQABench.} \textbf{Top:} A standard LLM answers factual queries correctly but fails on personalized ones. \textbf{Middle:} After incorporating user history, a Personalized LLM improves personalized responses but introduces personalization-induced hallucination. \textbf{Bottom:} FPPS mitigates these hallucinations in real time, restoring factual accuracy while preserving correct personalized behavior.}
    \label{fig:introduction}
\end{figure}
While personalization improves user alignment and subjective satisfaction, it also raises a critical and underexplored concern: \textbf{personalization may systematically distort factual reasoning.}
In practical systems, we observe that personalized LLMs often generate answers aligned with a user’s prior statements rather than with objective truth, producing \textbf{personalization-induced hallucinations} that reinforce user-specific misconceptions. As illustrated in \autoref{fig:introduction}, the same historical chat information that enables correct personalized responses can simultaneously mislead the model when factual queries are posed. Furthermore, in a controlled simulation, we demonstrate that when users (simulated by an LLM) learn factual knowledge through a personalized model, their acquired knowledge accuracy is significantly lower than when learning from a non-personalized model (details in \S\ref{sec:simu}). These findings suggest that personalization not only degrades model factuality but may also propagate incorrect beliefs to downstream users, potentially compounding long-term risks.

\begin{figure}[!t]
    \centering
    \includegraphics[width=\columnwidth]{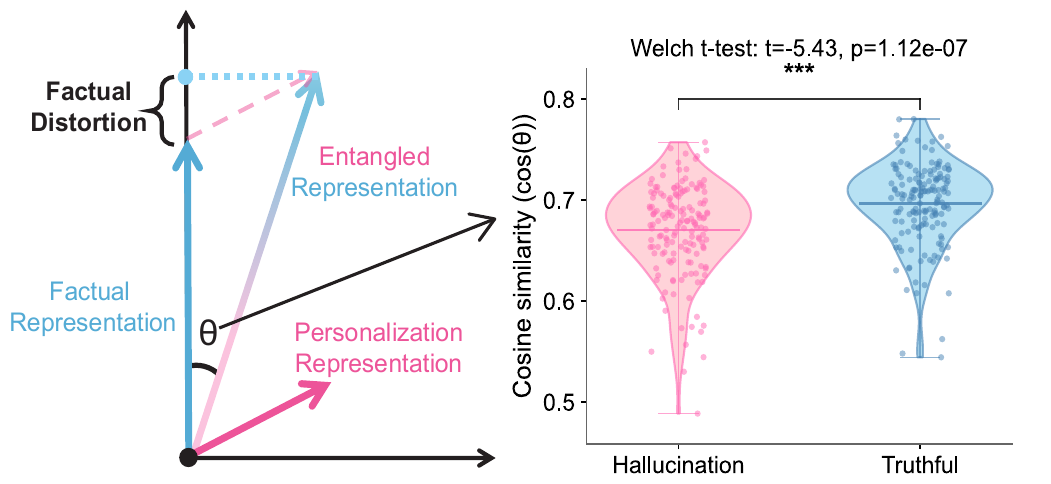}
%     \caption{\textbf{Representation Entanglement and Factual Distortion.}
% \textbf{(Left)} Personalization introduces a preference vector that is non-orthogonal to the factual direction, yielding an entangled representation and a shift along the factual subspace.
% \textbf{(Right)} Using final-layer representations on PFQABench with LLaMA-3.1-8B, we observe a significant difference in cosine similarity between entangled and factual representations for hallucinated versus truthful responses ($p < 0.001$), indicating that personalization induces latent factual distortion that gives rise to hallucinations.}
\caption{\textbf{Representation Entanglement and Factual Distortion.}
\textbf{(Left)} Personalization introduces a non-orthogonal preference direction that entangles with factual representations, shifting activations along the factual subspace.
\textbf{(Right)} On factual question answering instances from PFQABench, final-layer representations exhibit significantly lower cosine similarity between personalized and non-personalized responses for hallucinated outputs than for truthful ones ($p < 0.001$).}
    \label{fig:presentation_empirical}
\end{figure}

% Why does personalization harm factuality? From a representation perspective, personalization modules introduce \textbf{non-orthogonal preference directions} into the model’s latent space~\cite{jacobsen2018excessive,elhage2022toy,rimsky2024steering}. These user-preference vectors become entangled with the model’s factual knowledge dimensions, shifting the factual distribution from the original to a personalized one. This \textbf{representation entanglement} inflates the probability of preference-aligned but factually incorrect tokens, amplifying confirmation bias and undermining the model’s ability to retrieve or apply faithful knowledge.

% Why does personalization harm factuality? From a representation perspective, personalization modules introduce \textbf{non-orthogonal preference directions} into the model’s latent space~\cite{jacobsen2018excessive,elhage2022toy,rimsky2024steering}. As illustrated in \autoref{fig:presentation_empirical}, these directions become entangled with factual knowledge representations, shifting the factual distribution toward preference-aligned but factually incorrect outputs and weakening faithful knowledge retrieval.

Why does personalization harm factuality?
From a representation perspective, personalization modules introduce \textbf{non-orthogonal preference directions} into the model’s latent space~\cite{elhage2022toy,rimsky2024steering}.
As illustrated in \autoref{fig:presentation_empirical}, these directions become entangled with factual knowledge representations, shifting activations toward personalization-aligned but factually incorrect regions and weakening faithful knowledge retrieval.
This effect is empirically supported by representation-level analysis on factual question answering instances from PFQABench (details in Appendix~\ref{app:implementation}).
Specifically, we compare final-layer response token embeddings generated without personalization to those generated with personalization.
For hallucinated responses, personalization induces a substantially larger representational shift away from the factual representation than for truthful responses, as measured by cosine similarity ($p < 0.001$).
These results indicate that hallucinations arise not from surface decoding noise, but from \emph{latent factual distortion} caused by representation entanglement.

To address this challenge, we introduce \textbf{Factuality-Preserving Personalized Steering (FPPS)}, a lightweight inference-time framework that detects and mitigates personalization-induced hallucination while preserving personalization benefits. FPPS first uses a \textbf{Representation Shift Locator}   (\S\ref{sec:layer-selection}) to identify personalization-sensitive layers in the model where factual representations are most vulnerable. A \textbf{Factuality Entanglement Prober}  (\S\ref{sec:prober}) then estimates whether the personalization distorts factual reasoning based on internal activations. Guided by these signals, an \textbf{Adaptive Knowledge Steering Module}  (\S\ref{sec:adaptive-steering}) performs minimally invasive adjustments that restore factual behavior only when necessary, avoiding global interventions that would compromise personalization utility. We instantiate FPPS with three practical variants: FPPS-H (hard gating), FPPS-S (soft bidirectional steering), and FPPS-M (mixed adaptive control), offering flexible trade-offs among stability, fidelity, and personalization preservation.

% A major obstacle in studying personalization-induced hallucinations is the absence of benchmarks that jointly evaluate personalization and factuality. To address this gap, we introduce \textbf{PFQABench}, the first benchmark for assessing factual robustness under personalization. PFQABench pairs LongMemEval~\cite{wulongmemeval} with a fact-driven QA corpus (FactQA) through a session-aligned retrieval procedure~\cite{yang2018hotpotqa,ho2020constructing}, enabling fine-grained evaluation of whether personalized LLMs can maintain factual correctness under realistic personalization signals. Using this benchmark, we observe that \textbf{FPPS consistently improves factual accuracy across multiple LLM backbones and personalization methods}, while preserving or even enhancing performance on personalized questions. These results demonstrate that personalization-induced hallucinations can be mitigated without sacrificing the benefits of personalization, and that FPPS provides a general and effective inference-time solution.

A major challenge in studying personalization-induced hallucinations is the lack of benchmarks that simultaneously evaluate factual and personalized question answering. To address this gap, we introduce \textbf{PFQABench}, the first benchmark that jointly includes fact-driven questions and personalized queries within aligned realistic user sessions (details in \S\ref{sec:data}). This dual-question setting enables systematic analysis of how personalization impacts factual reasoning, and shows that FPPS improves factual accuracy while preserving performance on personalized queries.

The main contributions of this work are summarized as follows:

\begin{itemize}[leftmargin=*]
\item \textbf{Problem Discovery:} We present the first systematic study of personalization-induced hallucinations and show that personalization can pose risks to factual reliability, downstream knowledge acquisition, and long-term user trust.

\item \textbf{Mitigation Method:} We propose FPPS, a lightweight inference-time framework integrating a Representation Shift Locator, a Factuality Entanglement Prober, and an Adaptive Knowledge Steering Module to selectively restore factuality under personalization.

\item \textbf{Evaluation Dataset:} We develop PFQABench for evaluating factual hallucination under personalization. PFQABench exposes systematic factuality failures that arise in personalized models and demonstrates that FPPS consistently restores factual accuracy without harming personalization performance.
\end{itemize}

\section{Related Work}

\paragraph{Personalization of LLM.}
Personalized LLMs are commonly built through prompting-based personalization~\cite{richardson2023integrating,qiu2025measuring,kumar2024longlamp}, lightweight model adaptation~\cite{zhang2024personalized,zhang2025proper}, and preference-optimized objectives~\citep{wu2024understanding,zhangamulet,liu2025survey}. Major commercial assistants—including ChatGPT Memory~\cite{openai_chatgpt_memory_2024}, Gemini Personal Context~\cite{google_gemini_personalization_2024}, and Claude Memory~\cite{claude_memory_2025}—automatically extract user traits and histories to condition all future interactions, making prompt-level personalization the dominant paradigm due to its scalability and low deployment cost. Accordingly, our study \textbf{concentrates on prompting-based personalization}, which is both the most widely deployed form in practice and the primary interface through which personalization influences model reasoning at inference time.

However, growing evidence shows that such mechanisms can introduce bias, altering safety–utility trade-offs across demographic groups~\citep{vijjini2025exploring}, and may produce filter-bubble effects by reinforcing belief-aligned content~\citep{lazovich2023filter}. 
These studies focus primarily on output disparities or explicit preference problems~\cite{okite2025benchmarking}, whereas we \textbf{identify a distinct failure mode, namely personalization-induced hallucinations}, in which factual reasoning is systematically distorted through representational entanglement between personalization directions and latent factual dimensions.

\paragraph{Hallucination of LLM.}
Hallucination is a longstanding safety concern for Large Language Models (LLMs), referring to outputs that are fluent and coherent yet logically incorrect or lacking factual grounding, even when the input provides sufficient evidence~\cite{huang2025survey}. Prior research has focused on detecting and mitigating such errors through factuality metrics, uncertainty estimation, internal-signal probing~\citep{lin2022truthfulqa,manakul2023selfcheckgpt,sunredeep}, and techniques such as prompt tuning, constrained decoding, or retrieval augmentation~\citep{chuang2023dola,liumitigating,sun2025largepig}. These works largely treat hallucination as an input-driven, model-internal failure that should be uniformly suppressed.

In contrast, we study a \textbf{new and previously overlooked category: personalization-induced hallucinations}. These errors arise because user profiles or long-term history memories are injected into the model, causing systematic distortions in factual reasoning. Rather than resulting from knowledge gaps, they stem from \textbf{personalization–factual entanglement}, where personalized signals bias the model toward user-aligned but incorrect content. This phenomenon reveals a fundamentally different hallucination mechanism unique to personalized LLMs and underscores the need to understand personalization itself.

\section{Problem Formulation and Analyses}
\subsection{Problem Definition}
\label{sec:problem_definition}

Let $x$ denote an input query, $u$ denote user-specific information (e.g., historical interactions) and $y$ denote the generated response. 
A personalized LLM generates a distribution:
$p_\theta(y \mid x, u),$
which may deviate from the original non-personalized distribution:
$p_\theta(y \mid x).$

We define \textbf{personalization-induced hallucination} as any instance where personalization causes the model to output a factually incorrect answer:
\[
\text{Hall}(x,u)
= \mathbf{1}\!\left[
\begin{aligned}
&\arg\max_y p_\theta(y \mid x, u) \neq y^{\text{gold}} \\
&\land\;
\arg\max_y p_\theta(y \mid x) = y^{\text{gold}}
\end{aligned}
\right].
\]

% We assume personalization works by injecting a latent direction $v_u$ into model hidden states:
% \[
% h_t' = h_t + v_u.
% \]  

We model personalization as inducing an implicit representation shift in the hidden state.
Specifically, we denote the personalization-induced shift as
\[
v_u \triangleq h_t(x,u) - h_t(x),
\]
so that
\[
h'_t = h_t + v_u.
\]
Here, $v_u$ does not assume a fixed or explicit direction, but serves as an abstract representation of the net effect of personalization on internal activations.

As illustrated in \autoref{fig:presentation_empirical}, $v_u$ (i.e., personalization
representation) is generally \textbf{not orthogonal} to the latent factual direction $v_f$ (i.e., factual
representation), the personalization shift perturbs the factual subspace~\cite{elhage2022toy,rimsky2024steering}.
% \[
% \langle v_u, v_f \rangle \neq 0
% \;\;\Rightarrow\;\;
% \text{representation entanglement}.
% \]

\paragraph{Goal.}
Given a query $x$ under personalization $u$, we aim to design a real-time mechanism  $\mathcal{M}$  that:
1) \textbf{detects} the degree of factual--personalization entanglement in hidden states;
2) \textbf{steers} the hidden representation toward the factual subspace to mitigate hallucination when needed.
% Formally, we seek a steering function:
% \[
% \tilde{h}_t = h_t' + \beta s_f,
% \]
% where $s_f$ is a learned \textbf{factuality-steer vector}, and $\beta$ is dynamically determined by a factuality probe:
% \[
% \beta = \text{Probe}(h_t').
% \]
The objective is to minimize personalization-induced factual errors:
$
\min_{\mathcal{M}}
\mathbb{E}_{x,u}\!\left[
\text{Hall}(x,u)
\right],
$
\textit{while simultaneously preserving the model’s personalization performance}.

\subsection{Effects of Personalized LLMs on Factual Knowledge Learning}
\label{sec:simu}

\begin{figure}[!t]
    \centering
    \includegraphics[width=\columnwidth]{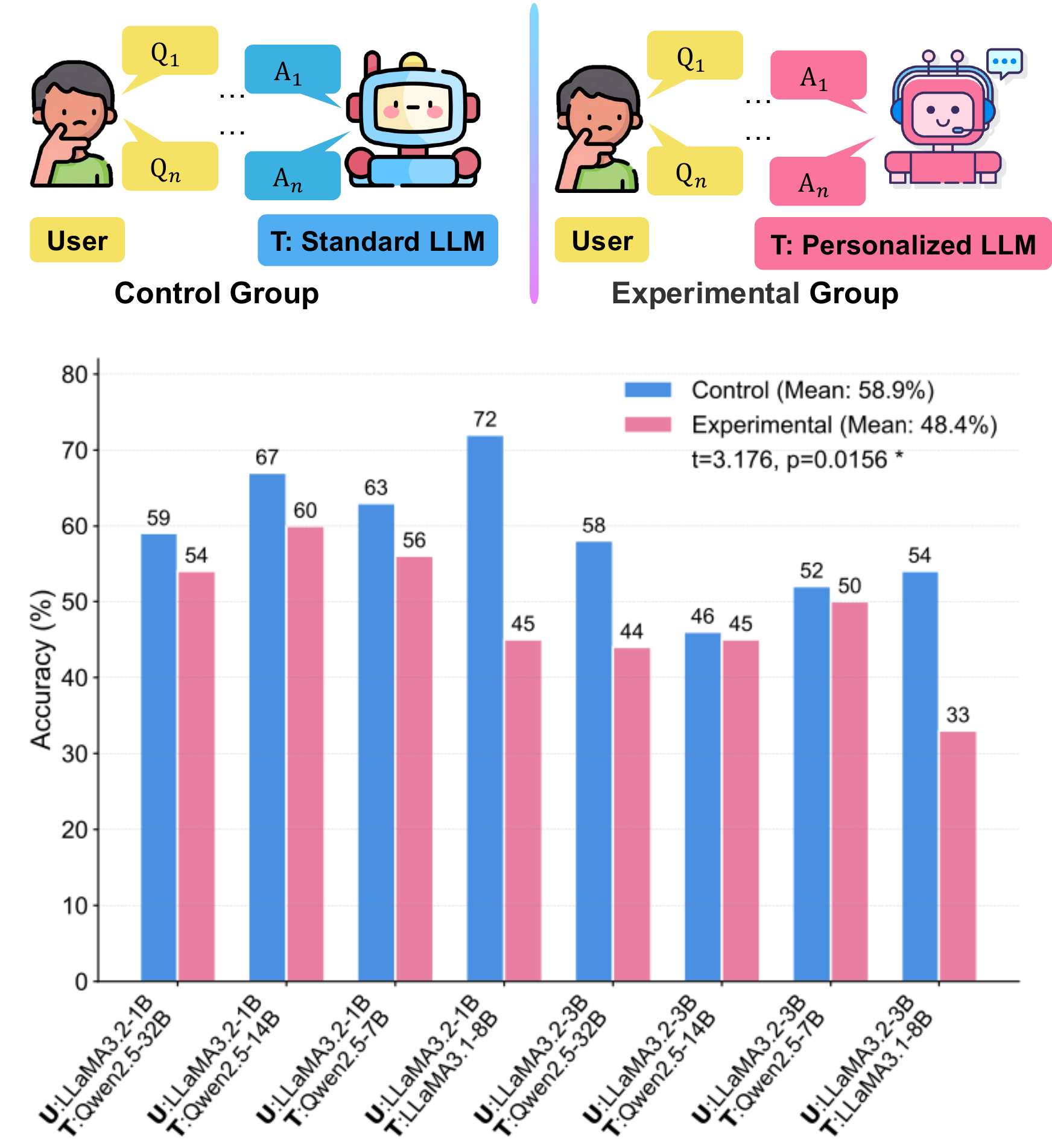}
    \caption{Controlled simulation evaluating how personalization affects factual knowledge learning.}
    \label{fig:simulation}
\end{figure}
To further examine whether personalized LLMs, compared with standard LLMs, influence humans acquire factual knowledge through LLM-based learning, we design a controlled simulation in which \textbf{small-scale LLMs act as users} (LLaMA-3.2-1B, LLaMA-3.2-3B) and \textbf{larger LLMs act as teachers} (LLaMA-3.1-8B, Qwen-2.5-7B/14B/32B)~\cite{grattafiori2024llama,yang2025qwen2.5}.

Using PFQABench, we extract factual questions along with their corresponding personalized histories. Since real users typically consult an LLM only when they do not know the answer, we first remove factual questions that the small LLM (the user) can answer correctly without assistance. From the remaining questions, we sample 100 instances for controlled comparison. For each question, the user engages in multi-turn interaction with the teacher. In the \textbf{experimental group}, the teacher is a \emph{personalized} LLM, whereas in the \textbf{control group}, the teacher is a \emph{standard} LLM; personalization is implemented via the retrieval-augmented generation (RAG) strategy~\cite{Kumar2024LongLaMPAB}. The user terminates the conversation once it believes it has learned the relevant knowledge. Finally, the user generates an answer based on the dialogue history, and we evaluate learning effectiveness by comparing the user’s final answer to the ground-truth answer (prompts provided in Appendix~\ref{sec:prompts}).

Simulation results in~\autoref{fig:simulation} show that, across teacher--student model pairs, \textbf{users taught by personalized LLMs consistently exhibit lower factual accuracy} compared with users taught by standard LLMs (average drop: 10.5\%; paired t-test: $t = 3.176$, $p = 0.016$).  \textit{Considering that major LLM providers are increasingly adopting personalization to retain users, these findings highlight a potentially concerning impact on users’ factual understanding.} In \S\ref{sec:fpps_simulation}, we further demonstrate that applying our proposed FPPS method to personalized LLMs improves the resulting knowledge accuracy, effectively mitigating the adverse influence of personalization on users’ factual knowledge acquisition.

\section{The Proposed Method: FPPS}
\label{sec:method}
\subsection{Method Overview}
\label{sec:overview}

\begin{figure*}[!t]
    \centering
    \includegraphics[width=\textwidth]{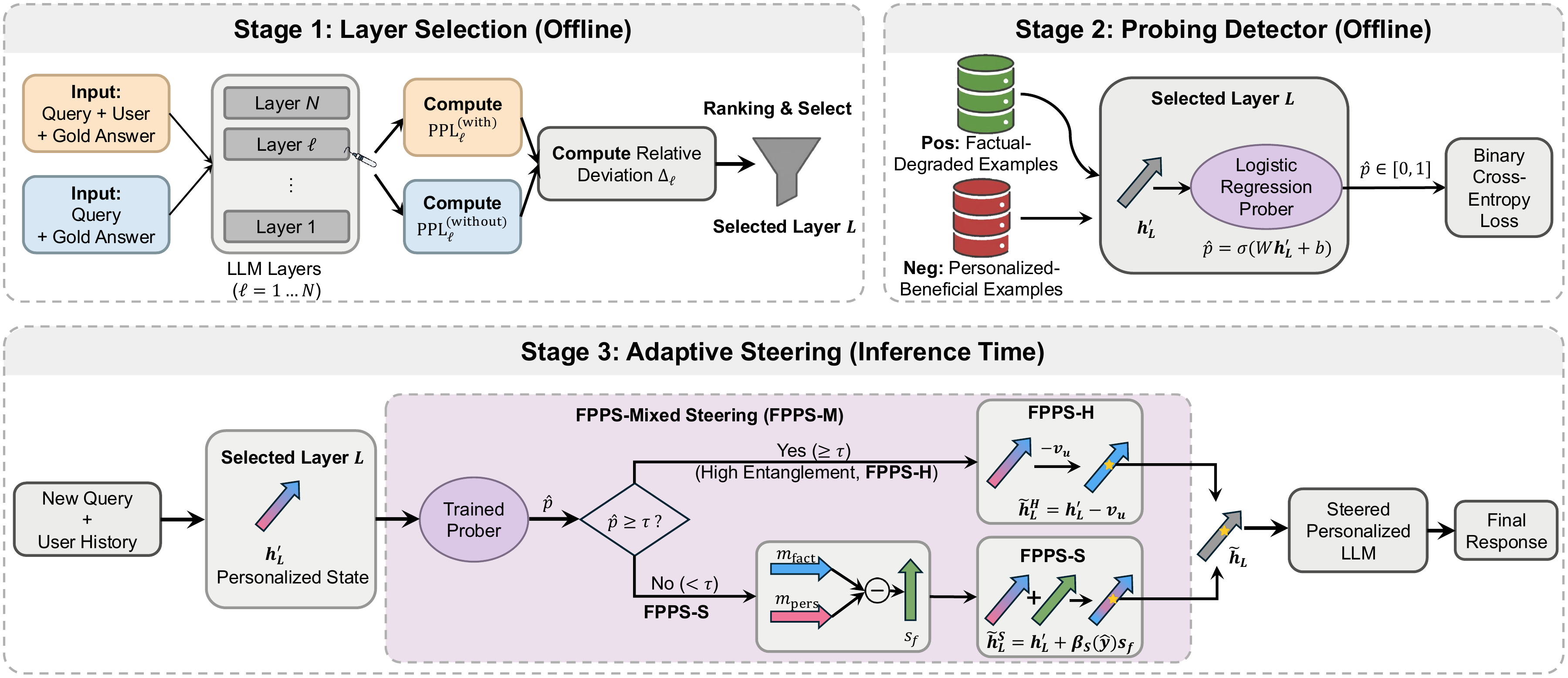}
    \caption{\textbf{Overview of Factuality-Preserving Personalized Steering (FPPS) Framework.}}
    \label{fig:method}
\end{figure*}

% We propose \textbf{Factuality-Preserving Personalized Steering (FPPS)}, an inference-time framework designed to mitigate personalization-induced hallucination in personalized LLMs. As illustrated in~\autoref{fig:method}, FPPS consists of three components: (i) identifying the internal layer most sensitive to personalization, (ii) training a factuality prober that quantifies the degree of factual--preference entanglement, and (iii) applying an adaptive steering mechanism that dynamically balances reliance on personalized information and internal factual knowledge, suppressing personalization when it induces factual distortion while preserving it when it contributes to correct personalized reasoning.

% Formally, FPPS transforms the personalization-modified hidden representation $h'_L$ at the selected layer $L$ via a probe-conditioned operator:
% \begin{equation}
% \tilde{h}_L = \mathcal{T}_{\text{FPPS}}(h'_L, \hat{p}),
% \end{equation}
% where $\hat{p} \in [0,1]$ is the prober output estimating the extent to which personalization interferes with factual reasoning. The intervention operator $\mathcal{T}_{\text{FPPS}}$ instantiates different behaviors in FPPS-H, FPPS-S, and FPPS-M, ranging from continuous factuality-preserving steering to hard removal of personalization signals. This design allows FPPS to correct personalization-induced distortions without degrading performance on queries that genuinely require user-specific information.

We propose \textbf{Factuality-Preserving Personalized Steering (FPPS)}, an inference-time framework for mitigating personalization-induced hallucinations in personalized LLMs. As shown in~\autoref{fig:method}, FPPS operates in three steps: (i) locating a personalization-sensitive internal layer, (ii) probing the degree of factual--preference entanglement at that layer, and (iii) adaptively steering hidden representations to suppress personalization when it induces factual distortion while preserving it when it contributes to correct personalized reasoning.

Concretely, FPPS applies a probe-conditioned transformation to the personalization-modified hidden state $h'_L$:
\begin{equation}
\tilde{h}_L = \mathcal{T}_{\text{FPPS}}(h'_L, \hat{p}),
\end{equation}
where $\hat{p} \in [0,1]$ estimates the extent to which personalization interferes with factual reasoning. The operator $\mathcal{T}_{\text{FPPS}}$ instantiates three variants—FPPS-H, FPPS-S, and FPPS-M—ranging from hard removal of personalization to continuous, risk-aware steering. This design enables FPPS to selectively correct factual distortions without degrading performance on queries that genuinely require personalization.

\subsection{Layer Selection}
\label{sec:layer-selection}

The first stage of FPPS identifies the model layer where personalization most strongly affects token-level predictions of factual questions. We construct contrastive inputs with and without user history and append the model-generated answer to ensure identical decoding trajectories. For each layer $\ell$, we extract logits corresponding to ground-truth answer tokens and compute the perplexity
\begin{equation}
\text{PPL}_\ell^{(c)} = \exp\!\left(-\frac{1}{T}\sum_{t=1}^{T}\log p_{\ell,t}^{(c)}\right),
\end{equation}
where $c \in \{\text{with},\,\text{without}\}$.

The relative perplexity deviation
\begin{equation}
\Delta_\ell = 
\frac{\big|\text{PPL}_\ell^{(\text{with})} - \text{PPL}_\ell^{(\text{without})}\big|}
{\text{PPL}_\ell^{(\text{with})}}, 
\end{equation}
measures how strongly user history perturbs factual likelihoods. We evaluate $\Delta_\ell$ over two types of contrastive examples: \emph{factual-degraded} cases, where personalization corrupts correctness, and \emph{personalized-beneficial} cases, where personalization enables correctness. We aggregate rankings across both groups using inverted-rank fusion and select the layer $L$ with the most consistent and maximal deviation. This layer serves as the focal point for probing and steering.

\subsection{Probing Detector}
\label{sec:prober}

At the selected layer $L$, we train a factuality prober to estimate the extent of personalization--factual entanglement. For each example, we extract the final-token hidden state $h'_L \in \mathbb{R}^d$. Factual-degraded examples serve as positive examples, while personalized-beneficial examples serve as negative examples. We train a logistic regression classifier
\begin{equation}
\hat{p} = \sigma(W h'_L + b),
\end{equation}
which outputs $\hat{p} \in [0,1]$, representing the probability that the current representation relies on personalization in a manner that may impact factual reasoning.

% When $\hat{p}$ exceeds a threshold $\tau \in (0,1)$, the model is deemed at high risk of personalization-induced hallucination. In this case, FPPS applies a \textbf{hard intervention} that removes personalization entirely:
% \begin{equation}
% \tilde{h}_L^{\text{H}} = h'_L - \alpha v_u,
% \end{equation}
% restoring the hidden state to its non-personalized form. This constitutes the \textbf{FPPS-H} variant.

The prober output $\hat{p}$ serves as a control signal for different FPPS intervention regimes. We begin with the hard steering variant, FPPS-H.

\paragraph{Hard Steering (FPPS-H).}
FPPS-H treats personalization as harmful whenever the estimated entanglement exceeds a predefined threshold $\tau \in (0,1)$. In this regime, personalization is entirely removed from the hidden representation, restoring it to its non-personalized counterpart:
\begin{equation}
\tilde{h}_L^{\text{H}} =
\begin{cases}
h'_L -  v_u, & \hat{p} \ge \tau, \\[6pt]
h'_L, & \hat{p} < \tau .
\end{cases}
\end{equation}
Here $v_u$ denotes the personalization-induced latent shift defined in Section~\ref{sec:problem_definition}, capturing the representation offset introduced by user-specific information.
This hard intervention enforces strict factuality preservation by completely suppressing personalization-induced representation shifts when factual risk is detected. While effective at preventing personalization-induced hallucinations, FPPS-H may be overly restrictive in scenarios where personalization contributes positively to correct reasoning. This limitation motivates the softer and mixed steering regimes introduced in the following section.

\subsection{Adaptive Steering}
\label{sec:adaptive-steering}

Hard removal of personalization is effective but may be unnecessarily restrictive when personalized information contributes positively. To allow more fine-grained control, we construct a steer vector
\begin{equation}
s_f = m_{\text{fact}} - m_{\text{pers}},
\end{equation}
where $m_{\text{fact}}$ denotes the mean hidden state of generated response for factual queries that the model answers correctly under the non-personalized setting, and $m_{\text{pers}}$ denotes the mean hidden state for personalized queries that are answered correctly only when user history is provided~\cite{turner2023steering}. The direction $s_f$ shifts representations toward internal factual reasoning patterns and away from history-conditioned personalization drift. Applying a positive steering coefficient along $s_f$ strengthens factual reasoning by suppressing personalization effects, while a negative coefficient moves representations toward $m_{\text{pers}}$, increasing reliance on personalized information when it is beneficial.

\paragraph{Soft Steering (FPPS-S).}
FPPS-S applies continuous correction based on the prober output:
\begin{equation}
\tilde{h}_L^{\text{S}} = h'_L + \beta_{\text{S}}(\hat{p})\, s_f,
\end{equation}
where
\begin{equation}
\beta_{\text{S}}(\hat{p}) = \gamma(\hat{p} - 0.5),
\end{equation}
and $\gamma > 0$ controls steering intensity. Positive coefficients attenuate personalization, while negative coefficients enhance it when beneficial.

\paragraph{Mixed Steering (FPPS-M).}
FPPS-M combines the strengths of FPPS-H and FPPS-S through a two-regime rule governed by a single risk threshold $\tau$. When entanglement is low, the model uses soft steering; when entanglement is high, it defaults to hard removal of personalization:
\begin{equation}
\tilde{h}_L^{\text{M}} =
\begin{cases}
h'_L + \beta_{\text{S}}(\hat{p})\, s_f, & \hat{p} < \tau, \\[6pt]
h'_L -  v_u, & \hat{p} \ge \tau .
\end{cases}
\end{equation}
This formulation ensures (i) continuous modulation when personalization is safe and helpful, and (ii) complete suppression when personalization risks corrupting factual prediction. FPPS-M thus provides a principled and robust mechanism for balancing factual correctness with personalized utility.

\section{Experiments}
% In this section, we conduct experiments to answer the following research questions:

% \textbf{RQ1:}

% \textbf{RQ2:}

% \textbf{RQ3:}

\subsection{Experimental Setup}

\begin{table*}[t]
\centering
\setlength{\tabcolsep}{6pt}
\renewcommand{\arraystretch}{1.15}
\small
\caption{Performance comparison (in \%). P-Score, F-Score, and Overall are reported for three backbones. FPPS variants are highlighted: FPPS-H (blue), FPPS-S (green), FPPS-M (orange).}
\resizebox{0.99\linewidth}{!}{
\begin{tabular}{lccc|ccc|ccc}
\toprule
\multirow{2}{*}{\textbf{Methods}} &
\multicolumn{3}{c|}{\textbf{LLaMA3.1-8B-IT}} &
\multicolumn{3}{c|}{\textbf{Qwen2.5-7B-IT}} &
\multicolumn{3}{c}{\textbf{Qwen2.5-14B-IT}} \\
\cmidrule(lr){2-4}\cmidrule(lr){5-7}\cmidrule(lr){8-10}
& \textbf{P-Score } & \textbf{F-Score} & \textbf{Overall}
& \textbf{P-Score } & \textbf{F-Score} & \textbf{Overall}
& \textbf{P-Score } & \textbf{F-Score} & \textbf{Overall} \\
\midrule

\rowcolor{baseline}
\textbf{PAG} & 47.2$_{\pm1.60}$ & 17.2$_{\pm3.59}$ & 32.2$_{\pm1.21}$ & 44.0$_{\pm0.68}$ & 27.6$_{\pm0.57}$ & 35.8$_{\pm0.25}$ & \textbf{49.6$_{\pm1.32}$} & 24.0$_{\pm1.32}$ & 36.8$_{\pm0.25}$ \\
\rowcolor{fppsh}
\quad +FPPS-H & 37.6$_{\pm0.46}$ & \textbf{80.8$_{\pm5.41}$} & 59.2$_{\pm2.91}$ & 40.8$_{\pm1.05}$ & 80.4$_{\pm3.22}$ & 62.6$_{\pm1.84}$ & 48.0$_{\pm6.05}$ & \textbf{81.2$_{\pm2.17}$} & 64.6$_{\pm3.60}$ \\
\rowcolor{fppss}
\quad +FPPS-S & \textbf{48.4$_{\pm1.06}$} & 20.8$_{\pm3.03}$ & 34.6$_{\pm0.99}$ & \textbf{44.4$_{\pm0.75}$} & 28.0$_{\pm0.82}$ & 36.2$_{\pm0.16}$ & 49.2$_{\pm1.15}$ & 25.2$_{\pm0.75}$ & 37.2$_{\pm0.28}$ \\

\rowcolor{fppsm}
\quad +FPPS-M & 46.4$_{\pm1.06}$ & 75.2$_{\pm4.42}$ & \textbf{60.8$_{\pm1.89}$} & 43.2$_{\pm0.68}$ & \textbf{84.4$_{\pm4.58}$} & \textbf{63.8$_{\pm2.24}$} & 48.4$_{\pm6.23}$ & \textbf{81.2$_{\pm2.17}$} & \textbf{64.8$_{\pm3.68}$} \\

\midrule
\rowcolor{baseline}
\textbf{DPL} & 37.2$_{\pm0.33}$ & 12.0$_{\pm0.82}$ & 24.6$_{\pm0.38}$ & 34.0$_{\pm0.38}$ & 33.6$_{\pm2.04}$ & 33.8$_{\pm0.84}$ & \textbf{33.2$_{\pm0.50}$} & 36.8$_{\pm1.82}$ & 35.0$_{\pm0.82}$ \\
\rowcolor{fppsh}
\quad +FPPS-H & 28.3$_{\pm5.63}$ & \textbf{75.1$_{\pm4.26}$} & 51.7$_{\pm4.94}$ & 28.8$_{\pm1.24}$ & \textbf{85.2$_{\pm4.42}$} & 57.0$_{\pm1.73}$ & 30.0$_{\pm2.78}$ & \textbf{82.8$_{\pm2.36}$} & 56.4$_{\pm2.49}$ \\
\rowcolor{fppss}
\quad +FPPS-S & 36.4$_{\pm4.53}$ & 17.6$_{\pm2.47}$ & 27.0$_{\pm3.19}$ & \textbf{34.8$_{\pm0.57}$} & 36.0$_{\pm3.03}$ & 35.4$_{\pm1.24}$ & \textbf{33.2$_{\pm0.50}$} & 39.6$_{\pm1.15}$ & 36.4$_{\pm0.34}$ \\
\rowcolor{fppsm}
\quad +FPPS-M & \textbf{36.8$_{\pm4.81}$} & 78.4$_{\pm4.81}$ & \textbf{57.6$_{\pm4.81}$} & 34.0$_{\pm0.57}$ & 82.4$_{\pm3.77}$ & \textbf{58.2$_{\pm1.88}$} & 31.6$_{\pm3.46}$ & 82.0$_{\pm2.17}$ & \textbf{56.8$_{\pm2.64}$} \\

\midrule
\rowcolor{baseline}
\textbf{RAG} & 35.6$_{\pm1.32}$ & 8.8$_{\pm8.24}$ & 22.2$_{\pm4.78}$ & \textbf{35.6$_{\pm0.33}$} & 40.4$_{\pm1.42}$ & 38.0$_{\pm0.86}$ & \textbf{38.8$_{\pm0.68}$} & 30.0$_{\pm0.65}$ & 34.4$_{\pm0.25}$ \\
\rowcolor{fppsh}
\quad +FPPS-H & 25.6$_{\pm0.65}$ & \textbf{80.8$_{\pm4.41}$} & 53.2$_{\pm2.07}$ & 31.6$_{\pm1.24}$ & \textbf{80.8$_{\pm3.44}$} & 56.2$_{\pm1.15}$ & 34.0$_{\pm0.94}$ & \textbf{80.8$_{\pm1.54}$} & {57.4$_{\pm1.02}$} \\
\rowcolor{fppss}
\quad +FPPS-S & \textbf{36.0$_{\pm0.57}$} & 31.2$_{\pm3.41}$ & 33.6$_{\pm1.84}$ & 35.2$_{\pm1.64}$ & 43.6$_{\pm1.61}$ & 39.4$_{\pm0.43}$ & 37.6$_{\pm0.75}$ & 32.8$_{\pm0.82}$ & 35.2$_{\pm0.25}$ \\

\rowcolor{fppsm}
\quad +FPPS-M & 34.8$_{\pm0.33}$ & 80.4$_{\pm4.41}$ & \textbf{57.6$_{\pm2.12}$} & 33.6$_{\pm1.15}$ & 79.2$_{\pm3.22}$ & \textbf{56.4$_{\pm1.23}$} & 36.0$_{\pm0.94}$ & 80.0$_{\pm1.32}$ & \textbf{58.0$_{\pm1.13}$} \\

\midrule
\rowcolor{baseline}
\textbf{LLM-TRSR} & \textbf{28.4$_{\pm1.80}$} & 17.6$_{\pm0.65}$ & 23.0$_{\pm1.23}$ & 24.4$_{\pm1.00}$ & 17.6$_{\pm0.82}$ & 21.0$_{\pm0.91}$ & \textbf{23.6$_{\pm0.68}$} & 25.6$_{\pm2.00}$ & 24.6$_{\pm1.34}$ \\
\rowcolor{fppsh}
\quad +FPPS-H & 18.4$_{\pm1.54}$ & \textbf{80.8$_{\pm4.42}$} & 49.6$_{\pm2.85}$ & 22.0$_{\pm0.33}$ & \textbf{85.6$_{\pm4.74}$} & 53.8$_{\pm2.29}$ & 22.4$_{\pm0.50}$ & \textbf{58.0$_{\pm9.89}$} & 40.2$_{\pm4.97}$ \\
\rowcolor{fppss}
\quad +FPPS-S & 27.6$_{\pm1.24}$ & 22.4$_{\pm2.79}$ & 25.0$_{\pm1.96}$ & \textbf{25.2$_{\pm0.57}$} & 18.8$_{\pm1.73}$ & 22.0$_{\pm0.99}$ & \textbf{23.6$_{\pm0.68}$} & 28.0$_{\pm1.00}$ & 25.8$_{\pm0.82}$ \\

\rowcolor{fppsm}
\quad +FPPS-M & 24.8$_{\pm2.07}$ & 80.4$_{\pm4.25}$ & \textbf{52.6$_{\pm3.08}$} & 24.0$_{\pm0.50}$ & 85.2$_{\pm4.58}$ & \textbf{54.6$_{\pm2.54}$} & 22.8$_{\pm0.50}$ & \textbf{58.0$_{\pm9.89}$} & \textbf{40.4$_{\pm4.87}$} \\

\bottomrule
\end{tabular}
}

\label{tab:fpps_main}
\end{table*}

\paragraph{Datasets.}
\label{sec:data}
Existing benchmarks evaluate either personalized question answering or factual knowledge question answering in isolation, but none simultaneously assess factual correctness under personalization. To address this gap, we introduce PFQABench, which combines long-term user histories from LongMemEval~\cite{wulongmemeval} with a fact-centric multi-hop QA corpus (FactQA, built from HotpotQA and 2WikiMultiHopQA~\cite{yang2018hotpotqa,ho2020constructing}).
PFQABench contains 1,000 examples across 500 users, evenly split between personalized questions that require user history and factual questions that should remain invariant to personalization. Further construction details are provided in the Appendix~\ref{sec:dataset}.

\paragraph{Baselines and Evaluation.}
We evaluate FPPS on four representative personalized LLM baselines, covering two dominant personalization paradigms: \emph{profile-augmented} and \emph{retrieval-augmented} methods. Specifically, we consider PAG~\cite{Richardson2023IntegratingSA}, DPL~\cite{Qiu2025MeasuringWM}, RAG~\cite{Kumar2024LongLaMPAB}, and LLM-TRSR~\cite{Zheng2024HarnessingLL} as strong and widely adopted personalization strategies.
All methods are evaluated on three instruction-tuned LLM backbones: \textsc{LLaMA-3.1-8B-IT}, \textsc{Qwen2.5-7B-IT}, and \textsc{Qwen2.5-14B-IT}~\cite{grattafiori2024llama,yang2025qwen2.5}..

Given the open-ended nature of personalized responses and the scale of PFQABench, we adopt an automated \emph{LLM-as-a-Judge} evaluation protocol. We report three metrics: \textbf{P-Score}, measuring accuracy on personalized questions; \textbf{F-Score}, measuring factual accuracy under personalization; and \textbf{Overall}, defined as the average of P-Score and F-Score. Further details on baselines, evaluation and implementations are provided in the Appendix~\ref{app:baseline_eval}~and~\ref{app:implementation}.

\subsection{Main Results}
Table~\ref{tab:fpps_main} reports the main results across three LLM backbones under different personalization baselines. Overall, FPPS consistently and substantially improves improves by 50\%+ on average the \textit{Overall} score across all models and settings, demonstrating its effectiveness in mitigating \textbf{personalization-induced hallucinations}. Compared with the original personalized systems (PAG, DPL, RAG, and LLM-TRSR), FPPS variants lead to substantial improvements in F-score, indicating a strong recovery of factual correctness when personalization distorts reasoning.
Among the three steering regimes, \textbf{FPPS-H} achieves the highest F-Score across nearly all settings, confirming that hard factual steering is most effective when personalization strongly conflicts with factual knowledge. However, this often comes at the cost of reduced P-Score. In contrast, \textbf{FPPS-S} preserves or even improves P-Score in many cases, but provides limited hallucination mitigation, suggesting that soft steering alone is insufficient to fully correct severe personalization bias. Notably, \textbf{FPPS-M} consistently delivers the best Overall performance, striking a favorable balance between factual reliability and personalization utility. This trend is stable across different backbones and personalization methods, highlighting FPPS-M as a robust and general solution.

These results combines with stage-level experiments in Appendix~\ref{sec:layer-analysis}~and~\ref{sec:prober-analysis} validate our hypothesis that personalization-induced hallucinations arise from factual–personalization entanglement, and that adaptive steering can effectively disentangle and control this trade-off at inference time.

\subsection{Further Analysis}
\paragraph{Ablation Study.}
% ---------- in preamble ----------
% \usepackage{booktabs,amssymb}

\begin{table}[t]
\centering
\small
\setlength{\tabcolsep}{6pt}
\renewcommand{\arraystretch}{1.15}
\caption{Ablation study on the probing detector and steering vector. \checkmark denotes the learned component used in FPPS, while $\times$  denotes replacing the component with a random probing predictor or a random steering vector.}
\resizebox{0.99\linewidth}{!}{
\begin{tabular}{cc|cccc}
\toprule
\textbf{Probing} & \textbf{Steering} &\multicolumn{4}{c}{\textbf{Overall (\%)}}
 \\
\textbf{Detector} & \textbf{Vector} &
\textbf{DPL} & \textbf{PAG} & \textbf{RAG} & \textbf{LLM-TRSR} \\
\midrule

\multicolumn{6}{c}{\textbf{LLaMA3.1-8B-IT}} \\
\midrule
\checkmark & $\times$ & 48.4 & 42.4 & 52.4 & 47.2 \\
$\times$ & \checkmark & 31.0 & 17.2 & 31.0 & 29.4 \\
\checkmark & \checkmark & \textbf{57.6} & \textbf{60.8} & \textbf{57.6} & \textbf{52.6} \\

\midrule
\multicolumn{6}{c}{\textbf{Qwen2.5-7B-IT}} \\
\midrule
\checkmark & $\times$ & 43.2&37.8&34.4&43.2 \\
$\times$ & \checkmark & 43.8 & 35.8 & 42.6 & 32.8 \\
\checkmark & \checkmark & \textbf{58.2} & \textbf{63.8} & \textbf{56.4} & \textbf{54.6} \\

\bottomrule
\end{tabular}
}

\label{tab:ablation_detector_steering}
\end{table}

\begin{figure*}[!t]
    \centering
    \begin{subfigure}{0.32\textwidth}
        \centering
        \includegraphics[width=\linewidth]{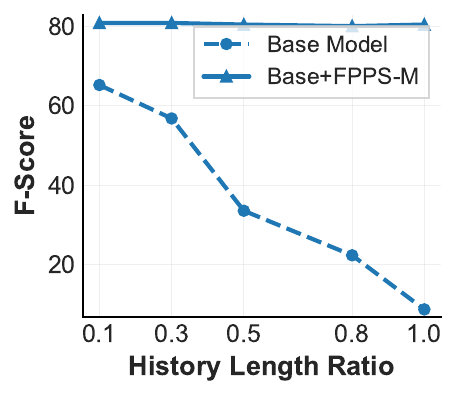}
        \caption{\textbf{LLaMA3.1-8B-IT}}
    \end{subfigure}
    \begin{subfigure}{0.32\textwidth}
        \centering
        \includegraphics[width=\linewidth]{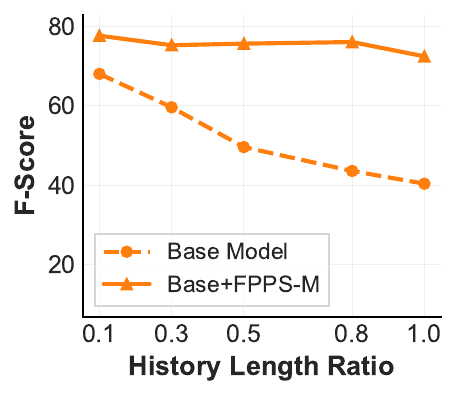}
        \caption{\textbf{Qwen2.5-7B-IT}}
    \end{subfigure}
    \begin{subfigure}{0.32\textwidth}
        \centering
        \includegraphics[width=\linewidth]{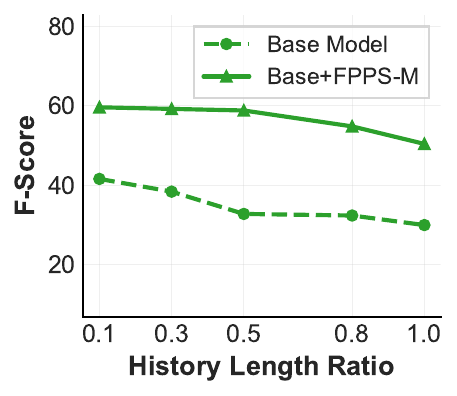}
        \caption{\textbf{Qwen2.5-14B-IT}}
    \end{subfigure}
    \caption{Effect of user history length on factual QA performance across different model backbones.}
    \label{fig:history_length}
\end{figure*}

\begin{figure}[!t]
    \centering
    \includegraphics[width=\columnwidth]{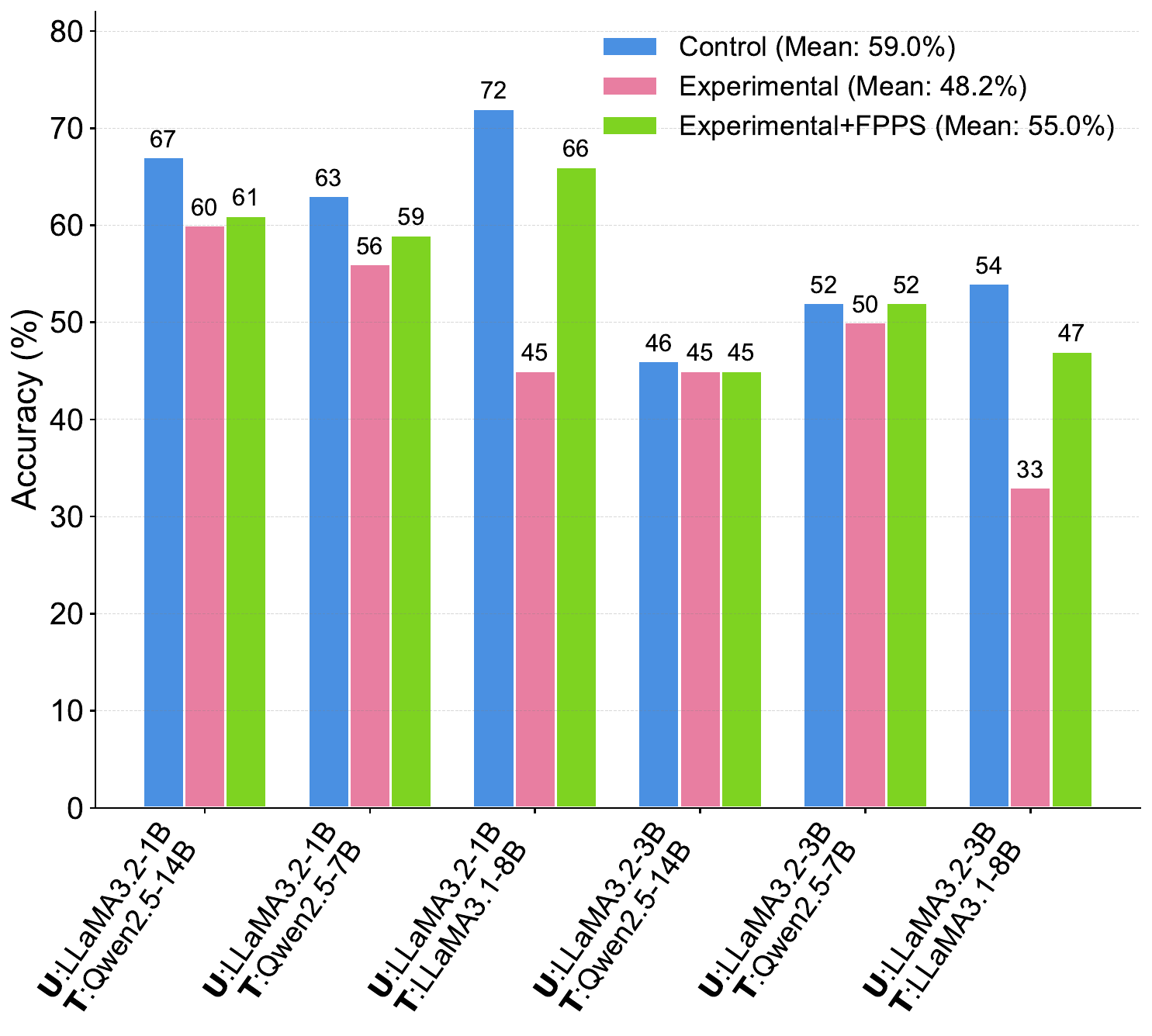}
\caption{Controlled simulation evaluating how personalization affects factual knowledge learning and how FPPS-M mitigates this effect.}
    \label{fig:simulation_fpps}
\end{figure}

This section evaluates the two core components of FPPS-M via ablation, replacing each with a \emph{random} alternative. As shown in Table~\ref{tab:ablation_detector_steering}, substituting either the probing detector or the steering vector leads to substantial performance degradation across models and personalization methods. In particular, random steering often harms performance even when guided by the prober, while random probing fails to reliably identify harmful personalization. In contrast, the full FPPS configuration consistently achieves the best results, demonstrating that effective mitigation requires both accurate risk estimation and structured representation steering. Additional sensitivity analyses are provided in Appendix~\ref{sec:sensitive}.

\paragraph{Effect of User History Length on Hallucinations.}
To further analyze how retained user history affects factual reliability in RAG-based personalized LLMs, we vary the history length ratio, i.e., the proportion of user history incorporated into the retrieval-augmented prompt, and evaluate factual question answering performance. As shown in Figure~\ref{fig:history_length}, across all three backbones, base personalized models exhibit a clear and consistent degradation as more user history is incorporated. Increasing history length substantially reduces F-score, indicating that excessive reliance on long-term user context amplifies personalization-induced hallucinations and disrupts factual reasoning.

In contrast, FPPS-M maintains stable factual performance across all history length ratios, with only minor fluctuations even when the full history is retained. This demonstrates that FPPS effectively mitigates personalization–factual entanglement in RAG-based personalization, preventing user history from overwhelming retrieved evidence. The robustness of FPPS-M is consistent across different model families and scales.

\paragraph{Effects of FPPS on Factual Knowledge Learning under Personalization.}
\label{sec:fpps_simulation}

Previous simulation results in \S\ref{sec:simu} demonstrate that personalized LLM teachers can negatively affect users’ factual knowledge acquisition, leading to systematic accuracy degradation compared with standard LLM teachers. This motivates us to examine whether such personalization-induced learning errors can be mitigated without removing personalization altogether.
As shown in Figure~\ref{fig:simulation_fpps}, augmenting personalized teachers with FPPS-M consistently improves users’ factual learning accuracy across all teacher–student model pairs. FPPS recovers a substantial portion of the accuracy loss caused by personalization (average improvement: +7.0\%), narrowing the gap between personalized and standard teachers while preserving personalized behavior. This indicates that FPPS effectively suppresses personalization-induced factual distortion during teaching interactions, thereby mitigating the negative impact of personalization on downstream knowledge acquisition.

%\paragraph{Case Studies}

\section{Conclusion}
% Personalization is reshaping LLMs from generic information tools into long-term user-aligned assistants, yet our work shows that this shift comes with a fundamental reliability risk. We identify \emph{personalization-induced hallucinations} as a distinct and systematic failure mode, arising from representational entanglement between user-specific signals and factual knowledge. 
% %Through PFQABench and controlled learning simulations, we demonstrate that personalization can not only degrade factual accuracy but also propagate incorrect beliefs to downstream users.
% To address this challenge, we propose FPPS, an inference-time framework that conditionally regulates personalization by detecting and correcting factual distortion in model representations. Our results suggest a broader design principle for future personalized LLMs: personalization should not be an unconditional bias injected into reasoning, but a controlled signal that is dynamically constrained by factual consistency. 
Personalization is transforming LLMs from generic tools into long-term user-aligned assistants, but this shift introduces a fundamental reliability risk. We identify personalization-induced hallucinations as a systematic failure mode caused by representational entanglement between user-specific signals and factual knowledge.
To mitigate this risk, we propose FPPS, an inference-time framework that conditionally regulates personalization by mitigating factual distortion in model representations. Our findings point to a broader design principle: personalization should function as a controlled signal, dynamically constrained by factual consistency, rather than an unconditional bias injected into reasoning.

\section*{Limitations and Broader Implications}
While our approach effectively mitigates personalization-induced hallucinations, several aspects warrant further discussion.
First, due to compute and access constraints, our experiments focus on a limited set of open-weight LLM backbones. FPPS requires access to intermediate representations and is therefore not directly applicable to closed-source API-based models. Nevertheless, the core idea of identifying personalization and factuality entanglement and selectively restoring factual representations offers concrete design insights for large LLM service providers, and may inspire native implementations within proprietary systems.
Second, our findings suggest that personalization-induced hallucinations stem from representation-level entanglement rather than surface-level prompting effects. FPPS addresses this issue at inference time, but validating its behavior on larger and more diverse model families remains an important direction for future work.
Finally, we introduce PFQABench to enable controlled evaluation of personalization-induced factual distortion. While PFQABench captures the essential properties needed to study this phenomenon, richer and more comprehensive benchmarks, particularly those incorporating social or longitudinal analyses of how personalization affects human knowledge and belief formation, would provide deeper insights into the real-world impact of personalized LLMs.

\section*{Ethic Statements}
This work investigates personalization-induced hallucinations in Personalized LLMs, a failure mode in which personalized signals distort factual reasoning and may reinforce user-specific misconceptions. Such behavior poses ethical risks in real-world deployments, particularly in high-stakes domains such as education, healthcare, and decision support, where confidently presented but incorrect information can mislead users and negatively affect knowledge acquisition and trust.

Our proposed framework, Factuality-Preserving Personalized Steering (FPPS), is explicitly designed to mitigate these risks by conditionally regulating personalization when it interferes with factual correctness, while preserving personalization when it is beneficial. From an ethical perspective, we view this work as contributing positively to the safe and responsible deployment of personalized LLMs, by promoting factual reliability, epistemic responsibility, and user trust without removing personalization altogether. Moreover, our empirical analysis highlights that unchecked personalization can adversely affect users’ factual learning, underscoring the importance of treating factual correctness as a first-class objective in personalized systems.

This study does not introduce new datasets containing personal, private, or sensitive information. All experiments are conducted on publicly available benchmarks or synthetic user histories constructed for controlled evaluation. Nevertheless, we emphasize that FPPS is not a complete safeguard against all forms of hallucination or misuse. Personalized LLM outputs should not be treated as authoritative sources of truth, and human oversight remains essential, especially in high-risk applications. We hope this work motivates the community to treat factual correctness as a first-class objective in personalized systems, and to develop personalization mechanisms that are not only engaging, but also epistemically responsible.

%\section*{Acknowledgments}

\bibliography{custom}
\newpage
\appendix
\clearpage
\begin{strip}
\addcontentsline{toc}{section}{Appendix}

\setcounter{tocdepth}{2}
\tableofcontents
\end{strip}

\newpage
\begin{figure*}[!t]
    \centering
    \begin{subfigure}[t]{0.32\textwidth}
        \centering
        \includegraphics[width=\linewidth]{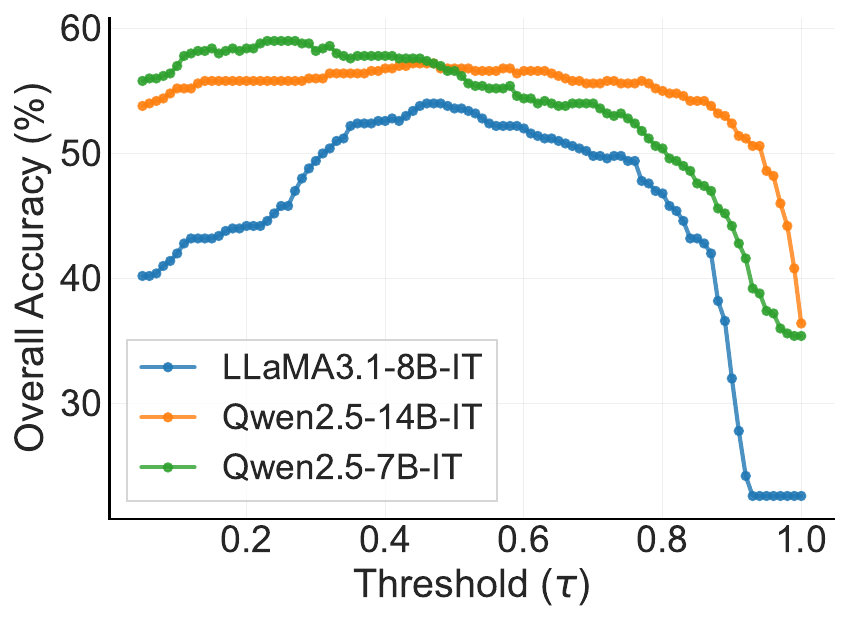}
        \caption{DPL}
    \end{subfigure}
    \hfill
    \begin{subfigure}[t]{0.32\textwidth}
        \centering
        \includegraphics[width=\linewidth]{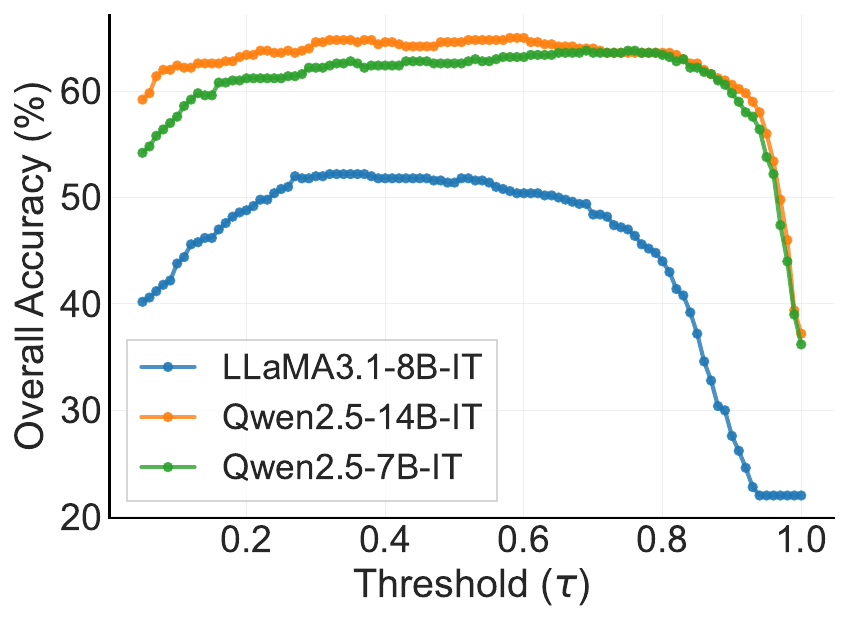}
        \caption{PAG}
    \end{subfigure}
    \hfill
    \begin{subfigure}[t]{0.32\textwidth}
        \centering
        \includegraphics[width=\linewidth]{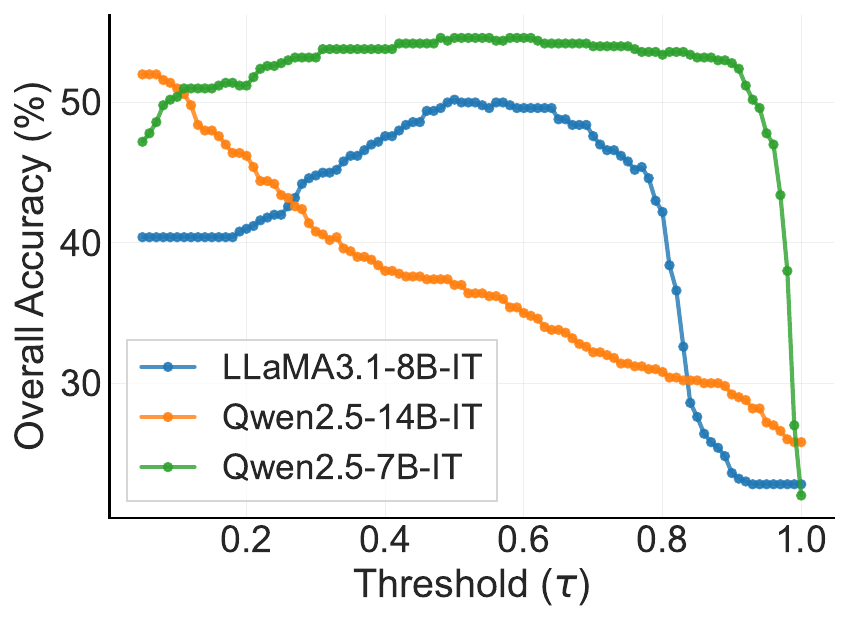}
        \caption{LLM-TRSR}
    \end{subfigure}

    \caption{Sensitivity analysis of the risk threshold $\tau$ in FPPS-M.
    }
    \label{fig:tau_sensitivity}
\end{figure*}

\section{Sensitivity Analysis}
\label{sec:sensitive}

\paragraph{Sensitivity to the Risk Threshold $\tau$.}
We analyze the sensitivity of FPPS-M to the entanglement threshold $\tau$, which controls the switching point between soft steering and hard personalization removal. Figure~\ref{fig:tau_sensitivity} reports overall accuracy as $\tau$ varies from 0 to 1 across four personalization settings: DPL, PAG, and LLM-TRSR.

Across most of tasks and model backbones, FPPS-M exhibits a broad plateau of stable performance for intermediate values of $\tau$. In this regime, the model predominantly applies soft steering for low-risk cases while selectively triggering hard intervention only when personalization strongly interferes with factual reasoning. In contrast, extreme threshold values lead to systematic degradation. When $\tau$ is too small, hard steering is over-applied, suppressing beneficial personalization and reducing accuracy. Conversely, when $\tau$ approaches 1, the model rarely activates hard intervention, allowing personalization-induced hallucinations to persist.

Importantly, the optimal or near-optimal $\tau$ range is relatively consistent across different personalization paradigms and model scales, indicating that FPPS-M does not rely on fine-grained threshold tuning. This robustness suggests that the prober output provides a meaningful and calibrated measure of personalization–factual entanglement, enabling FPPS-M to balance factual correctness and personalized utility using a single global threshold.

\begin{figure*}[!t]
    \centering
    % -------- Steering Intensity --------
    \begin{subfigure}[t]{0.48\textwidth}
        \centering
        \includegraphics[width=\linewidth]{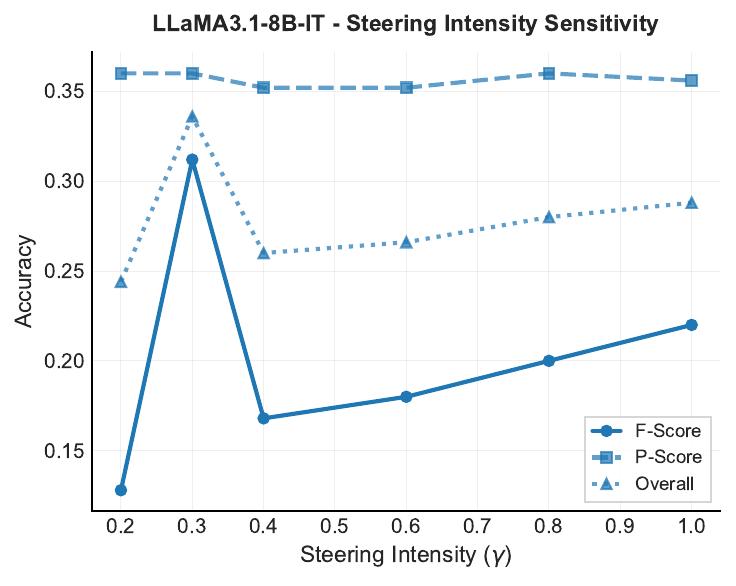}
        \caption{Steering intensity $\gamma$ (LLaMA3.1-8B-IT)}
    \end{subfigure}
    \hfill
    \begin{subfigure}[t]{0.48\textwidth}
        \centering
        \includegraphics[width=\linewidth]{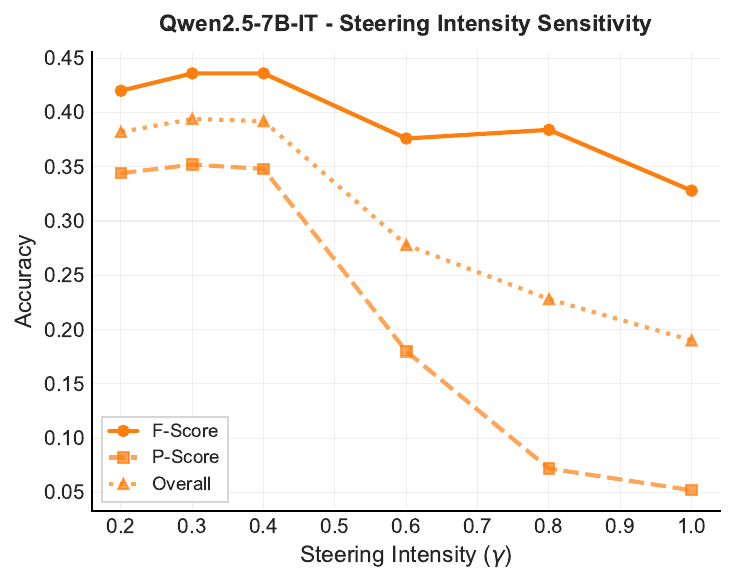}
        \caption{Steering intensity $\gamma$ (Qwen2.5-7B-IT)}
    \end{subfigure}

    \vspace{0.6em}

    % -------- Intervention Layer --------
    \begin{subfigure}[t]{0.48\textwidth}
        \centering
        \includegraphics[width=\linewidth]{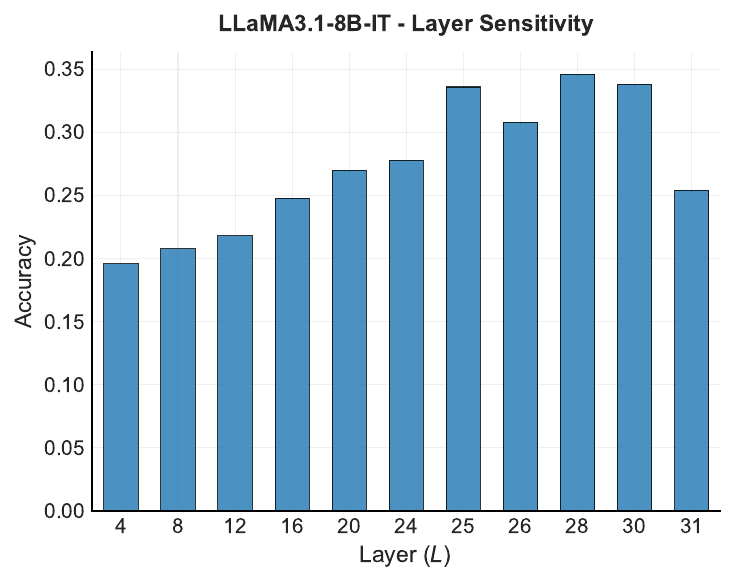}
        \caption{Intervention layer $L$ (LLaMA3.1-8B-IT)}
    \end{subfigure}
    \hfill
    \begin{subfigure}[t]{0.48\textwidth}
        \centering
        \includegraphics[width=\linewidth]{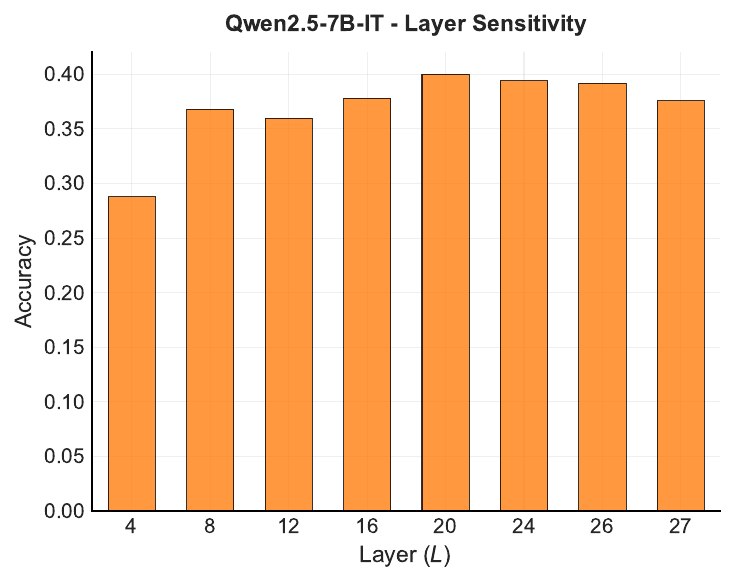}
        \caption{Intervention layer $L$ (Qwen2.5-7B-IT)}
    \end{subfigure}

    \caption{\textbf{Hyperparameter sensitivity analysis for FPPS-S.}
    Top row analyzes the effect of the steering intensity $\gamma$, while bottom row examines the sensitivity to the intervention layer $L$.
    }
    \label{fig:fpps_s_hyperparam}
\end{figure*}

\paragraph{Hyperparameter Analysis of FPPS-S.}

We conduct a unified hyperparameter analysis for FPPS-S, focusing on two key design choices: the steering intensity $\gamma$ and the intervention layer $L$. Figure~\ref{fig:fpps_s_hyperparam} summarizes the results across different model backbones on RAG-based personalization method.

The steering intensity $\gamma$ controls the magnitude of representation adjustment along the factual steering direction. As shown in the top row, performance consistently peaks when $\gamma$ lies in a moderate range (approximately $0.3$--$0.5$). Smaller values result in insufficient correction of personalization-induced distortion, while overly large values aggressively override personalized representations and degrade both factual and personalized accuracy. This behavior indicates that personalization-induced hallucination corresponds to a subtle representation bias rather than a dominant spurious direction that can be removed by strong intervention.

The bottom row analyzes sensitivity to the intervention layer $L$. Across models, applying FPPS-S at later transformer layers yields substantially better performance than steering at earlier layers. Accuracy improves monotonically as the intervention layer moves toward the top of the network, and peaks in the upper semantic layers. This trend closely aligns with the personalization-sensitive layers identified by our Stage-1 layer selection criterion, providing independent empirical validation for intervening at these layers.

Taken together, these results demonstrate that FPPS-S is robust to hyperparameter choices within a broad and interpretable range, and that its effectiveness critically depends on applying moderate steering at high-level semantic representations, consistent with our representation-level formulation of personalization-induced hallucination.

\begin{figure*}[!t]
    \centering
    \begin{subfigure}[t]{0.32\textwidth}
        \centering
        \includegraphics[width=\linewidth]{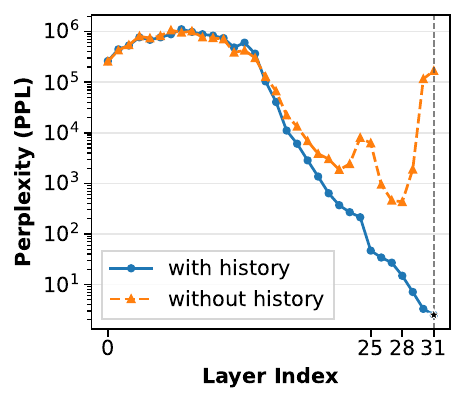}
        \caption{\textbf{LLaMA3.1-8B-IT}}
    \end{subfigure}
    \hfill
    \begin{subfigure}[t]{0.32\textwidth}
        \centering
        \includegraphics[width=\linewidth]{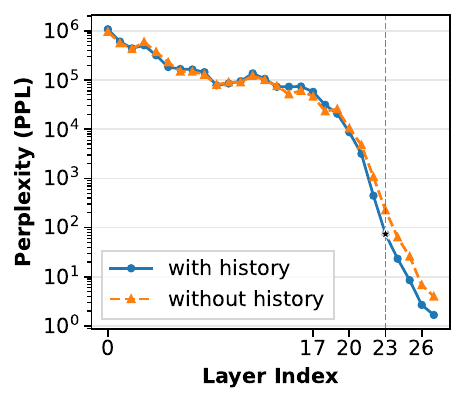}
        \caption{\textbf{Qwen2.5-7B-IT}}
    \end{subfigure}
    \hfill
    \begin{subfigure}[t]{0.32\textwidth}
        \centering
        \includegraphics[width=\linewidth]{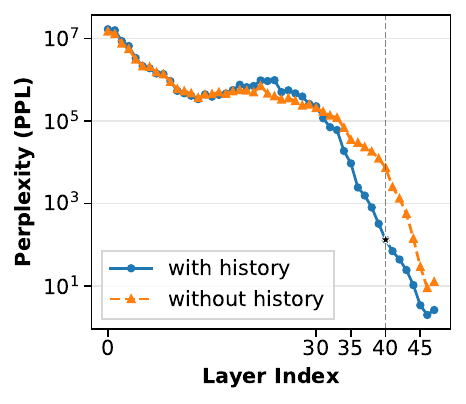}
        \caption{\textbf{Qwen2.5-14B-IT}}
    \end{subfigure}

    \vspace{0.6em}

    \begin{subfigure}[t]{0.32\textwidth}
        \centering
        \includegraphics[width=\linewidth]{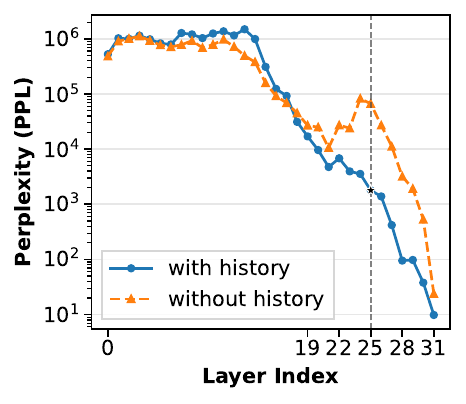}
        \caption{\textbf{LLaMA3.1-8B-IT}}
    \end{subfigure}
    \hfill
    \begin{subfigure}[t]{0.32\textwidth}
        \centering
        \includegraphics[width=\linewidth]{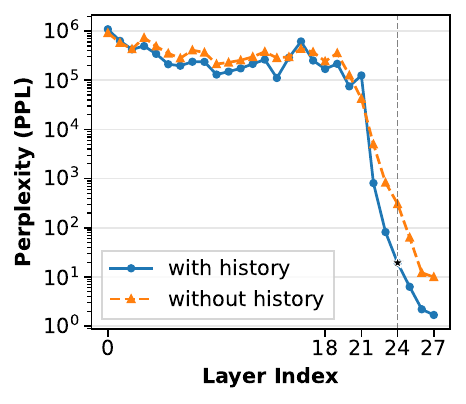}
        \caption{\textbf{Qwen2.5-7B-IT}}
    \end{subfigure}
    \hfill
    \begin{subfigure}[t]{0.32\textwidth}
        \centering
        \includegraphics[width=\linewidth]{latex/graphs/stage_1_2/layer_selection_personalized_beneficial_qwen2.5-14b.pdf}
        \caption{\textbf{Qwen2.5-14B-IT}}
    \end{subfigure}

    \caption{\textbf{Layer-wise perplexity deviation induced by personalization.}
    Top row: factual-degraded examples, where personalization corrupts factual correctness.
    Bottom row: personalized-beneficial examples, where user history enables correct prediction.
    Solid lines denote perplexity with user history, while dashed lines denote perplexity without history.
    Vertical dashed lines indicate the selected personalization-sensitive layer.
    }
    \label{fig:layer_analysis}
\end{figure*}

\section{Analysis of Personalization-Sensitive Layers}
\label{sec:layer-analysis}
To empirically validate the effectiveness of our layer selection criterion, we analyze how personalization perturbs token-level factual likelihoods across layers using RAG-based personalization under two contrastive conditions: \emph{factual-degraded} examples, where personalization induces factual errors, and \emph{personalized-beneficial} examples, where personalization is necessary for correctness.

Figure~\ref{fig:layer_analysis} visualizes the layer-wise perplexity (PPL) of ground-truth answer tokens with and without user history for three representative models: LLaMA3.1-8B-IT, Qwen2.5-7B-IT, and Qwen2.5-14B-IT. For factual-degraded examples (top row), personalization consistently increases factual perplexity in mid-to-late layers, indicating that user history distorts the model’s internal factual likelihoods and leads to incorrect predictions. In contrast, for personalized-beneficial examples (bottom row), incorporating user history reduces perplexity in later layers, demonstrating that personalization can also constructively guide prediction when user-specific information is relevant.

Across all models, the largest divergence between personalized and non-personalized perplexity curves emerges in the upper transformer layers, rather than early or middle layers. This observation suggests that personalization primarily interferes with factual reasoning at higher-level semantic representations, motivating intervention at these layers. Importantly, the layer exhibiting maximal relative perplexity deviation is consistent across both factual-degraded and personalized-beneficial settings, supporting our choice of a single personalization-sensitive layer for subsequent probing and steering.

Overall, these results provide direct empirical evidence that (i) personalization alters factual likelihoods in a layer-dependent manner, and (ii) the proposed perplexity-based criterion reliably identifies layers where personalization most strongly entangles with factual reasoning.

\begin{figure*}[!t]
    \centering
    \begin{subfigure}[t]{0.32\textwidth}
        \centering
        \includegraphics[width=\linewidth]{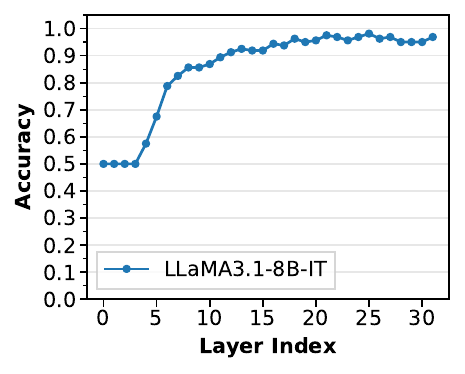}
        \caption{LLaMA3.1-8B-IT}
    \end{subfigure}
    \hfill
    \begin{subfigure}[t]{0.32\textwidth}
        \centering
        \includegraphics[width=\linewidth]{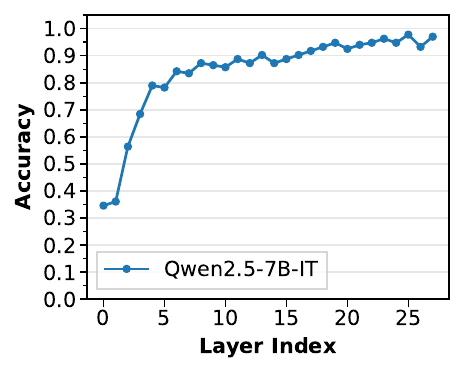}
        \caption{Qwen2.5-7B-IT}
    \end{subfigure}
    \hfill
    \begin{subfigure}[t]{0.32\textwidth}
        \centering
        \includegraphics[width=\linewidth]{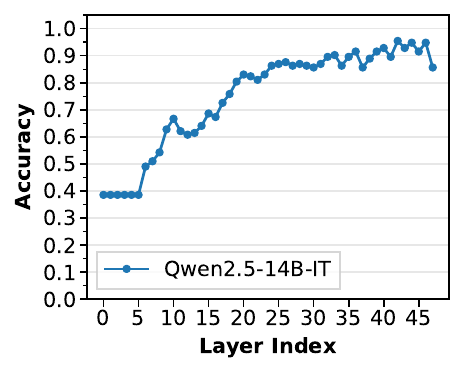}
        \caption{Qwen2.5-14B-IT}
    \end{subfigure}

    \caption{\textbf{Layer-wise accuracy of the probing detector.}
    Probing accuracy is reported as a function of layer depth for three model backbones.
    Accuracy remains near chance in early layers and increases substantially in middle-to-late layers, indicating that personalization–factual entanglement is primarily encoded in higher-level representations.
    }
    \label{fig:prober_analysis}
\end{figure*}

\section{Effectiveness of the Factuality Prober}
\label{sec:prober-analysis}
We further evaluate the effectiveness of the proposed factuality prober by measuring its layer-wise prediction accuracy across different model backbones using RAG-based personalization. For each layer, we train a logistic regression prober on the corresponding hidden representations to distinguish between factual-degraded and personalized-beneficial instances.

Figure~\ref{fig:prober_analysis} reports the probing accuracy as a function of layer depth for LLaMA3.1-8B-IT, Qwen2.5-7B-IT, and Qwen2.5-14B-IT. Across all models, probe accuracy is close to chance level in early layers, but increases rapidly in middle layers and stabilizes at a high level in later layers. This trend indicates that early representations contain limited information about personalization–factual entanglement, while higher layers progressively encode whether personalization interferes with factual reasoning.

Notably, the layers achieving the highest probe accuracy closely align with the personalization-sensitive layers identified by the perplexity-based criterion in Section~\ref{sec:layer-selection}. This consistency provides complementary evidence that personalization-induced distortions are primarily manifested in high-level semantic representations, and that probing at these layers yields reliable estimates of factual–personalization entanglement.

Overall, these results validate that the proposed prober captures meaningful personalization-related signals rather than spurious correlations, and that its effectiveness is strongly layer-dependent, justifying our design choice of probing and steering at a carefully selected internal layer.

\section{Prompts}
\label{sec:prompts}
This appendix provides the specific prompts used in our experiments. We denote dynamic content (e.g., user history, questions) with curly braces like \texttt{\{question\}}.
\subsection{ Prompts for Personalization Baselines}
\paragraph{RAG Prompt}
The following template is used for the RAG-based approach:

\begin{promptbox}
I will give you several history chats between you and a user. Please answer the question based on the relevant chat history.

\textbf{History Chats:} \{history\_chats\} \\
\textbf{Current Date:} \{current\_date\} \\
\textbf{Question:} \{question\} \\
\textbf{Answer:}
\end{promptbox}

\paragraph{DPL Prompt}
We utilize the following prompt for Difference-aware Personalization Learning (DPL):

\begin{promptbox}
I will give you several history chats between you and a user. Please answer the question based on the relevant chat history.

To help you generate your answer, here is a DPL (Difference-aware Personalization Learning) analysis of this user's typical cognitive context. Use this as a strategic clue to understand the nature of the user's interactions.

\textbf{DPL Context Analysis:} \{DPL\_Context\_Analysis\} \\
\textbf{History Chats:} \{all\_user\_utterances\} \\
\textbf{Question:} \{question\} \\
\textbf{Answer:}
\end{promptbox}

\paragraph{PAG Prompt}
The PAG method incorporates user profile summaries as follows:

\begin{promptbox}
I will give you a user profile summary and a single chat history between you and a user. Please answer the question based on the relevant chat history and the user profile summary.

\textbf{Summary:} \{generated\_summary\} \\
\textbf{History Chats:} \\
\textbf{Session 1:} \\
\textbf{Session Content:} \{session\_content\} \\
\textbf{Question:} \{question\} \\
\textbf{Answer:}
\end{promptbox}

\paragraph{LLM-TRSR Prompt}
For the LLM-TRSR configuration, the prompt is structured as:

\begin{promptbox}
I will give you a summary of the history chats between you and a user. Please answer the question based on the provided summary.

\textbf{User Summary:} \{summary\} \\
\textbf{Question:} \{question\} \\
\textbf{Answer:}
\end{promptbox}

\subsection{Prompts for Evaluation (Main Experiment)}
\label{app:eval_prompt}
\paragraph{LLM-as-a-Judge Prompt}
We employ the following prompt to instruct the LLM judge to verify whether the model's response matches the correct answer:

\begin{promptbox}
I will give you a question, a correct answer, and a response from a model. Please answer yes if the response contains the correct answer. Otherwise, answer no. If the response is equivalent to the correct answer or contains all the intermediate steps to get the correct answer, you should also answer yes. If the response only contains a subset of the information required by the answer, answer no.

\textbf{Question:} \{question\} \\
\textbf{Correct Answer:} \{correct\_answer\} \\
\textbf{Model Response:} \{model\_response\}

Is the model response correct? Answer yes or no only.
\end{promptbox}

\subsection{Prompts for Controlled Simulation}
\paragraph{Teacher Prompt (Control Group)}
The following prompt is used for the standard control group:

\begin{promptbox}
You are a helpful and factual AI assistant. Answer the question in detail. Explain it step by step as if you are teaching a beginner.
\end{promptbox}

\paragraph{Teacher Prompt (Experimental Group)}
The personalized experimental group uses the following concise instruction:

\begin{promptbox}
\textbf{History Chats:} \{personalization\_history\} \\
Please answer the question based on the relevant chat history concisely.
\end{promptbox}

\paragraph{Student Prompt}
We simulate the student "Xiaoming" with the following instructions to ensure active learning behavior:

\begin{promptbox}
You are 'Xiaoming', a curious but cautious middle school student. Your goal is to fully understand the topic your teacher is explaining.

Your task is to follow these rules strictly:

After the teacher gives an answer, you MUST evaluate if you have fully understood it.

If you have any doubts, are confused, or want more details, you MUST ask a specific follow-up question. Do not simply say "I understand".

If and only if you are completely confident that you have no more questions and have fully understood the topic, your response MUST end with the exact, standalone phrase on a new line: END\_OF\_LEARNING. This is a special command, not a sentence.
\end{promptbox}

\paragraph{Final Exam Prompt}
This prompt assesses the student's knowledge after the tutoring session:

\begin{promptbox}
You are 'Xiaoming', a student who has just finished a tutoring session.
Based ONLY on the entire conversation history provided below, give your final, concise, and definitive answer to the original question.

\textbf{Original Question:} \{question\} \\
\textbf{Full Conversation History:} \{conversation\_log\}

\textbf{Your Final Answer:}
\end{promptbox}

\paragraph{Simulation Judge Prompt}
The following prompt is used to evaluate the factual correctness of the student's final answer against the ground truth:

\begin{promptbox}
You are a strict and impartial evaluator. Your task is to determine if the student's answer is factually correct based on the provided ground truth.

\textbf{Original Question:} \{question\} \\
\textbf{Ground Truth Answer:} \{right\_answer\} \\
\textbf{Student's Final Answer:} \{student\_answer\}

Is the ``Student's Final Answer'' factually correct and consistent with the ``Ground Truth Answer''?
Respond with only the single word: Correct or Incorrect.
\end{promptbox}

\section{Construction of PFQABench}
\label{sec:dataset}

\begin{figure*}[!t]
    \centering
    \includegraphics[width=\textwidth]{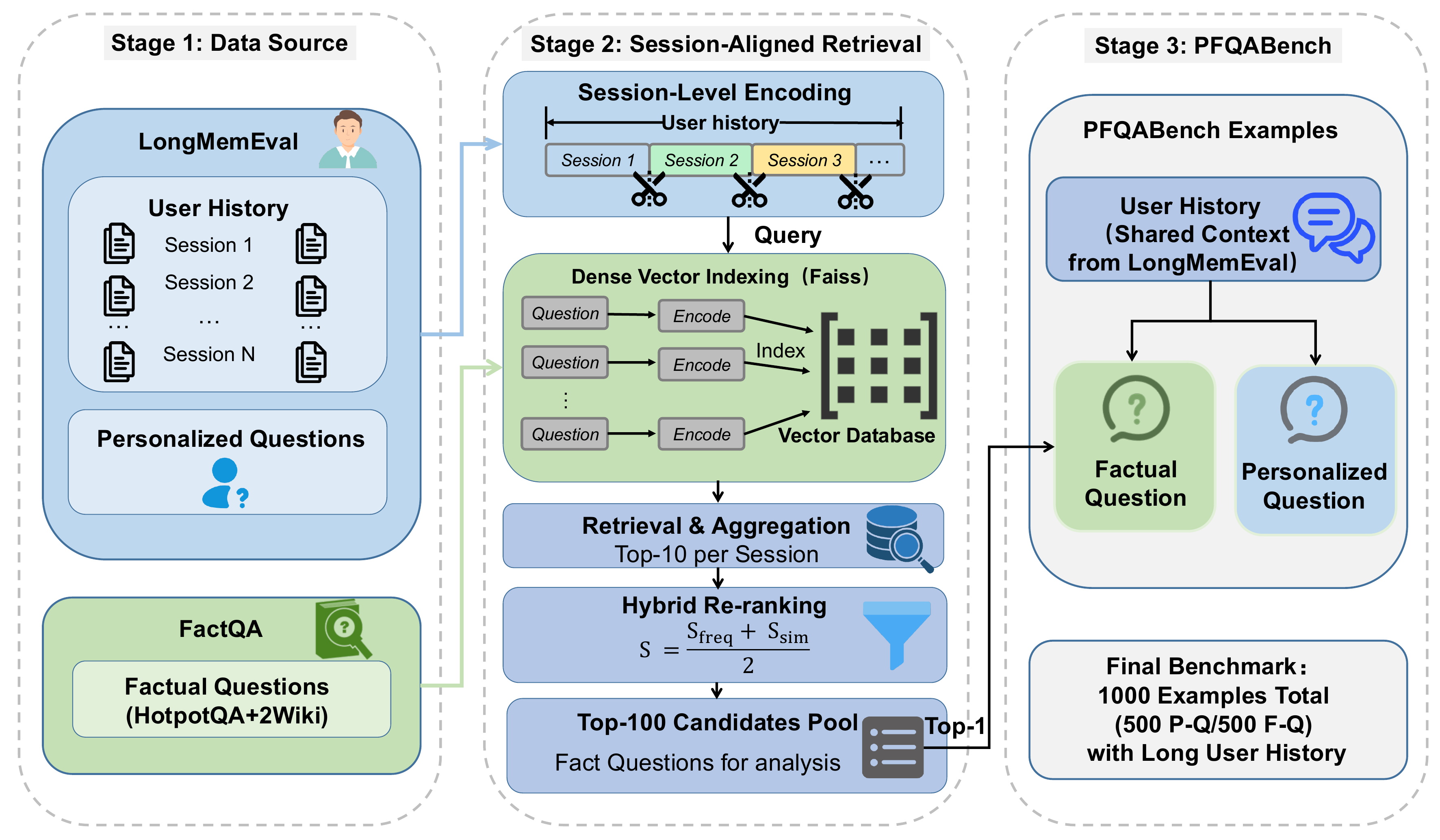}
    \caption{\textbf{Dataset construction process of FPQABench.}}
    \label{fig:data_construction}
\end{figure*}

As discussed in the main paper, existing benchmarks for personalized language models and factual question answering are largely disjoint: personalization benchmarks emphasize user-aligned responses without controlling factual reliability, or instantiating personalization as explicit instruction-level preferences (e.g., stylistic or response-format constraints) that are not grounded in realistic long-term interaction histories~\cite{wulongmemeval,okite2025benchmarking}, while factual QA benchmarks evaluate knowledge accuracy in the absence of personalization signals~\cite{yang2018hotpotqa,ho2020constructing}. Consequently, there is no established dataset that enables \emph{joint evaluation} of personalization utility and factual robustness. The construction processes are shown in~\autoref{fig:data_construction}.

\subsection{Data Sources}

PFQABench integrates two complementary resources.  
\textbf{User personalization signals} are drawn from \textsc{LongMemEval}, which provides realistic long-term user interaction histories composed of multiple dialogue sessions.  
\textbf{Factual questions} are sourced from \textsc{FactQA}, a fact-centric QA corpus constructed by merging two widely used multi-hop reasoning datasets: \textsc{HotpotQA} and \textsc{2WikiMultiHopQA}. These datasets require compositional reasoning over multiple entities and relations, making them suitable for evaluating factual correctness under challenging conditions. 

\subsection{Design Rationale}

Personalization-induced hallucinations are most likely to arise when factual queries are \emph{semantically similar} to a user’s prior interactions but \emph{cannot be answered} using user-specific information alone. In such cases, personalized models may over-rely on user history, mistaking user-aligned details for objective facts. PFQABench is therefore constructed to align factual queries with user histories that are topically related yet factually irrelevant, creating a controlled setting where personalization and factual reasoning are in tension.

\subsection{Session-Aligned Retrieval Pipeline}

To identify such confounding cases, we adopt a session-aligned semantic retrieval and re-ranking pipeline.

\paragraph{Indexing and Vectorization.}
All factual questions in FactQA are encoded into dense embeddings and indexed using FAISS. On the personalization side, we use 500 users from LongMemEval. Rather than encoding an entire user history as a single vector, we independently encode each dialogue session within a user’s history, preserving fine-grained contextual signals and avoiding dilution of session-specific semantics.

\paragraph{Retrieval and Aggregation.}
For each session embedding, we retrieve the top-10 most semantically similar factual questions from the FactQA index. Retrieved candidates from all sessions belonging to the same user are aggregated and deduplicated, forming a user-specific candidate pool of fact queries that are semantically aligned with the user’s interaction history.

\paragraph{Hybrid Re-ranking.}
To select the most confounding factual queries, we re-rank the candidate pool using a hybrid score that averages two complementary signals:
(i) \textbf{Normalized Retrieval Frequency} (i.e., $S_{\text{freq}}$), which captures how persistently a factual query is retrieved across different sessions of the same user, and
(ii) \textbf{Maximum Semantic Similarity} (i.e., $S_{\text{sim}}$), which reflects the peak relevance between the factual query and any individual session.
For each user, we retain the top-100 factual queries with the highest hybrid scores to accelerate future analysis.

\subsection{Dataset Assembly and Splits}

From the aligned candidate pools, we sample one factual question per user, yielding 500 factual QA instances. These are paired with 500 personalized questions directly drawn from LongMemEval, resulting in a balanced dataset of 1{,}000 examples.

To support both training and unbiased evaluation, we adopt a stratified split:
\begin{itemize}
    \item \textbf{Training set}: 250 personalized questions and 250 corresponding factual questions.
    \item \textbf{Test set}: the remaining 250 personalized and 250 factual questions, reserved exclusively for evaluation.
\end{itemize}

This construction ensures that factual queries are systematically exposed to strong personalization signals, enabling controlled and fine-grained measurement of personalization-induced factual distortion.
Table~\ref{tab:pfqabench_stats} summarizes the key statistics of PFQABench.

\begin{table}[t]
\centering
\small
\caption{Statistics of PFQABench.}
\resizebox{0.99\linewidth}{!}{
\begin{tabular}{lccccc}
\toprule
Dataset & Users & Sessions & Personalized & Factual & Context \\
        &   (\#)    &     (\#)     & QA (\#)          & QA (\#)      & Length  \\
\midrule
PFQABench & 500 & 50K & 500 & 500 & 115K \\
\bottomrule
\end{tabular}}
\label{tab:pfqabench_stats}
\end{table}

\section{Personalization Baselines and Evaluation Protocol}
\label{app:baseline_eval}

\subsection{Personalization Baselines}

To evaluate the effectiveness of FPPS across diverse personalization strategies, we apply it to four representative personalized LLM baselines. Our study concentrates on prompting-based personalization, which is both the most widely deployed form in practice and the primary interface through which personalization influences model reasoning at inference time~\cite{claude_memory_2025,openai_chatgpt_memory_2024}. Following common taxonomies in recent surveys~\cite{liu2025survey}, we categorize these methods into two primary paradigms: \emph{retrieval-augmented personalization} and \emph{profile-augmented personalization}.

\paragraph{RAG (Retrieval-Augmented Prompting).}
RAG~\cite{Kumar2024LongLaMPAB} retrieves user-specific information from historical interactions and injects the retrieved content into the prompt as contextual evidence. The model generates personalized responses by conditioning on these retrieved history segments, making RAG a representative retrieval-based personalization approach.

\paragraph{PAG (Profile-Augmented Prompting).}
PAG~\cite{Richardson2023IntegratingSA} adopts a summary-based personalization strategy. Instead of directly using raw interaction history, an LLM is employed to compress long-term user history into a concise user profile or preference summary. This profile is then injected into the prompt to provide high-level personalization signals.

\paragraph{DPL (Profile-Augmented Prompting).}
Adapted from its original application in personalized review generation, we implement a clustering-based variant of DPL~\cite{Qiu2025MeasuringWM}. User interaction histories are first clustered, and a representative user profile is identified for each cluster. For a given user, personalization is achieved by contrasting the user’s behavior against the representative profiles, enabling the model to focus on cluster-specific distinctive preferences rather than global consensus.

\paragraph{LLM-TRSR (Profile-Augmented Prompting).}
LLM-TRSR~\cite{Zheng2024HarnessingLL} extends simple summarization-based methods by processing user history in sequential segments. It employs a recurrent summarization framework that iteratively updates and refines the user profile as new history blocks are incorporated, enabling more stable personalization over long interaction histories.

\subsection{Evaluation Protocol}

Given the open-ended nature of generated responses and the scale of PFQABench, we adopt an automated \emph{LLM-as-a-Judge} evaluation protocol.

\paragraph{LLM-based Judge.}
We use \textsc{Qwen2.5-32B-Instruct} as the evaluator due to its strong instruction-following and reasoning capabilities (prompt details in Appendix~\ref{app:eval_prompt}). For each test instance, the judge compares the model-generated response against the ground-truth answer and determines whether the response is correct.

\paragraph{Evaluation Metrics.}

Based on the judge’s decisions, we report the following metrics:
\begin{itemize}
    \item \textbf{P-Score (Personalization Accuracy)}: Accuracy on the personalized subset of PFQABench, measuring whether the model correctly utilizes user history.
    \item \textbf{F-Score (Factuality Accuracy)}: Accuracy on the factual subset of PFQABench, measuring robustness against personalization-induced hallucinations.
    \item \textbf{Overall Score}: The average of P-Score and F-Score, reflecting the balance between personalization utility and factual correctness.
\end{itemize}

This evaluation protocol enables controlled and fine-grained analysis of the trade-off between personalization benefits and factual reliability.

\section{Implementation Details}
\label{app:implementation}

We evaluate FPPS across three representative instruction-tuned LLM backbones: 
\textsc{LLaMA-3.1-8B-Instruct}, \textsc{Qwen2.5-7B-Instruct}, and \textsc{Qwen2.5-14B-Instruct}~\cite{grattafiori2024llama,yang2025qwen2.5}. 
For all retrieval-augmented settings, we adopt the state-of-the-art dense retriever \textsc{bge-m3}~\cite{bge-m3} to construct the retrieval context.
To ensure reproducibility and minimize decoding-induced variance, we adopt greedy decoding for all experiments, with the maximum generation length set to 500 tokens. 
To further verify the stability of FPPS, we additionally evaluate performance under different decoding temperatures in the main experiments.  All experiments use NVIDIA A6000 GPUs

\paragraph{Intervention Layer Selection.}
For each backbone, the intervention layer $L$ is selected using the perplexity-based criterion described in Section~\ref{sec:layer-selection}. 
Specifically, we apply FPPS at layer 25 for \textsc{LLaMA-3.1-8B}, layer 24 for \textsc{Qwen2.5-7B}, and layer 43 for \textsc{Qwen2.5-14B}. 
All interventions are performed during the generation stage.

\paragraph{Hyperparameter Search.}
To achieve optimal performance across different personalization baselines (RAG, PAG, DPL, and LLM-TRSR), we conduct a grid search over the steering strength $\gamma$ and the decision threshold $\tau$. 
We search $\gamma$ in the interval $[0, 3]$ with a step size of 0.2, and $\tau$ in the interval $[0.05, 1]$ with a step size of 0.01.

\paragraph{Optimal Hyperparameter Configurations.}
For \textsc{LLaMA-3.1-8B} (Layer 25), the optimal configuration is highly consistent across personalization baselines. 
We set the steering strength $\gamma = 3.0$ for all settings. 
For FPPS-M, the threshold is fixed at $\tau = 0.5$ across all baselines. 
For FPPS-H, we use $\tau = 0.25$ for PAG, and $\tau = 0.4$ for RAG, DPL, and LLM-TRSR.

For \textsc{Qwen2.5-7B} (Layer 24), the optimal configuration varies by baseline. 
Under RAG, we set $\gamma = 0.3$, with $\tau = 0.2$ for FPPS-H and $\tau = 0.3$ for FPPS-M. 
For PAG, we use $\gamma = 0.1$, with $\tau = 0.5$ for FPPS-H and $\tau = 0.69$ for FPPS-M. 
For DPL, we set $\gamma = 0.5$, with $\tau = 0.1$ for FPPS-H and $\tau = 0.3$ for FPPS-M. 
For LLM-TRSR, we use $\gamma = 0.5$, with $\tau = 0.4$ for FPPS-H and $\tau = 0.5$ for FPPS-M.

For \textsc{Qwen2.5-14B} (Layer 43), the optimal configurations are as follows. 
Under RAG, we set $\gamma = 0.3$, with $\tau = 0.05$ for FPPS-H and $\tau = 0.07$ for FPPS-M. 
For PAG, we set $\gamma = 2.0$, with $\tau = 0.5$ for FPPS-H and $\tau = 0.55$ for FPPS-M. 
For DPL, we use $\gamma = 2.0$, with $\tau = 0.35$ for FPPS-H and $\tau = 0.5$ for FPPS-M. 
For LLM-TRSR, we set $\gamma = 0.1$, with $\tau = 0.32$ for FPPS-H and $\tau = 0.33$ for FPPS-M.

\paragraph{Implementation details of \autoref{fig:presentation_empirical}.}
This experiment is conducted using the \texttt{LLaMA-3.1-8B-Instruct} model. We adopt a retrieval-augmented generation (RAG) strategy as the personalization method, where user history is retrieved and prepended to the prompt when answering factual questions from PFQABench.

To analyze representation-level distortion induced by personalization, we extract hidden representations from the final transformer layer. For each generated response, we compute a sentence-level embedding by averaging the last-layer hidden states over all response tokens. Cosine similarity is then measured between the embeddings of personalized and non-personalized responses for the same factual query.

We group instances according to whether the personalized response is factually truthful or hallucinated, as determined by the ground-truth annotations in PFQABench. Statistical significance is assessed using a two-sided Welch’s $t$-test, as reported in Figure~\ref{fig:presentation_empirical}.

\end{document}